\documentclass{article}

\usepackage{arxiv}

\usepackage[utf8]{inputenc} % allow utf-8 input
\usepackage[T1]{fontenc}    % use 8-bit T1 fonts
\usepackage{hyperref}       % hyperlinks
\usepackage{url}            % simple URL typesetting
\usepackage{booktabs}       % professional-quality tables
\usepackage{amsfonts}       % blackboard math symbols
\usepackage{nicefrac}       % compact symbols for 1/2, etc.
\usepackage{microtype}      % microtypography
\usepackage{lipsum}
\usepackage{graphicx}
\graphicspath{ {./images/} }

\usepackage{blindtext}
\usepackage{amsmath}
\usepackage{amssymb}
\usepackage{amsfonts}
\usepackage{subcaption}
\usepackage{tikz}
\usepackage{tikz-cd}
\usepackage{tikz-3dplot}
\usepackage{algorithm}
\usepackage{algpseudocode}
\usepackage{multirow}

\newtheorem{theorem}{Theorem}

\newtheorem{conjecture}[theorem]{Conjecture}

\newtheorem{definition}[theorem]{Definition}

\newtheorem{remark}[theorem]{Remark}

\title{Prototype Selection using Topological Data Analysis (Preprint)}

\author{
 Jordan Eckert \\
  Department of Mathematics \& Statistics\\
  Auburn University\\
  Auburn, AL 36849 \\
  \texttt{jpe0018@auburn.edu} \\
   \And
 Elvan Ceyhan \\
  Department of Mathematics \& Statistics\\
  Auburn University\\
  Auburn, AL 36849 \\
  \texttt{ceyhan@auburn.edu} \\
  \And
 Henry Schenck \\
  Department of Mathematics \& Statistics\\
  Auburn University\\
  Auburn, AL 36849 \\
  \texttt{hks0015@auburn.edu} \\
}

\begin{document}
\maketitle
\begin{abstract}
Prototype selection methods compress a training set, but the existing taxonomy of condensation, edition, hybrid, competence-based, optimization-based, and clustering-based families does not include methods that operate on the multi-scale topological structure of the data. This paper introduces two different persistence-based prototype selector variants, Topological Prototype Selector (TPS) and Boundary-Conscious Topological Prototype Selector (BoundaryTPS). TPS uses two sequential Rips filtrations to retain boundary-relevant and interior-typical points. BoundaryTPS is a single-stage variant whose vertex-weighted filtration concentrates retention near the decision boundary. We evaluate both methods against seven classical baselines on fifteen real datasets and find that the topological methods occupy a different operating point in the prototype-selection design space than existing methods. BoundaryTPS achieves the lowest mean Friedman rank on $H_1$ persistence-diagram preservation and is significantly better than five of the seven baselines (Nemenyi, $\alpha = 0.05$). TPS ranks third on the same endpoint. Both methods are more stable under fold perturbation than any chained-decision selector tested, and both inherit the source set's class proportions without label-aware machinery. On aggregate G-Mean both methods are competitive but not leading, with rank-1 frequencies of $11.3\%$ (TPS) and $9.9\%$ (BoundaryTPS) across fold combinations. Empirically, both methods scale sub-quadratically in sample size. 
\end{abstract}

% keywords can be removed
\keywords{Data Reduction \and Prototype Selection \and Instance Selection \and Topological Data Analysis \and Persistent Homology}

% =============================================================================
% SECTION 1: Introduction
% =============================================================================
\section{Introduction}
Modern classification pipelines increasingly process datasets large enough that the size of the training set, not the cost of a single prediction, dominates inference time. The classical nearest-neighbor classifier illustrates this most directly as it stores the entire training set, incurs a per-query cost that scales linearly with that storage, and is sensitive to label noise within the retained instances~\cite{cover_nearest_1967, garcia_prototype_2012, wilson_asymptotic_1972}. Parametric alternatives such as support vector machines (SVM), random forests (RF), and multilayer perceptrons (MLP) shift the bottleneck from prediction to training time and memory, as both still scale with the number of training instances~\cite{bottou_support_2007, breiman_random_2001}. Prototype selection addresses both regimes by compressing the training set to a small representative subset, which serves as a standard preprocessing step in machine learning tasks~\cite{garcia_prototype_2012, olvera_lopez_prototype_2010}. The literature distinguishes two strategies for producing the compressed set. \textit{Instance selection} retains a subset of the original training instances~\cite{garcia_instance_2015}, while \textit{prototype generation} produces new synthetic instances, typically cluster centroids, that need not appear in the original data index~\cite{garcia_prototype_2012, olvera_lopez_prototype_2010}. We use \textit{prototype selection} throughout to refer to both strategies, following the convention of~\cite{garcia_prototype_2012}.

The prototype selection problem asks for a subset of the training data that is much smaller than the original yet still supports comparable downstream performance. We evaluate this trade-off along three axes. Structural fidelity is measured by comparing the persistence diagrams of the selected prototype set and the original training set, using the Wasserstein-1 distance between $H_1$ diagrams, the $L_2$ distance between $\beta_0$ and $\beta_1$ Betti curves, and the $L_2$ distance between Euler characteristic curves~\cite{cohen-steiner_lipschitz_2010, richardson_efficient_2014}. Selection stability is measured by the mean fold-to-fold Jaccard similarity of the selected prototype indices~\cite{kalousis_stability_2007}. Downstream classification quality is measured by the geometric mean of per-class recall (G-Mean) rather than raw accuracy, to remain informative under the class imbalance present across our benchmark datasets~\cite{japkowicz_assessment_2013}.

Classical methods span condensation, edition, hybrid, competence-based, optimization-based, and clustering-based families~\cite{garcia_prototype_2012, olvera_lopez_prototype_2010}, which we expand on further in Section~\ref{sec:prototype_selection_problem}. Despite the diversity of these families, each operates without an explicit representation of the data's topological structure. The geometry of the between-class decision boundary, which may exhibit complex structure across multiple spatial scales, is captured implicitly through these methods rather than through scale-resolved topological invariants.

Topological data analysis (TDA) offers the missing representation. Persistent homology summarizes a point cloud across a range of spatial scales, tracking connected components and loops (and higher-dimensional analogues), ultimately separating topological features that persist under scale change from those that do not~\cite{carlsson_topology_2009, edelsbrunner_computational_2010}. The framework has proven useful in supervised learning through vectorized persistence representations such as persistence images~\cite{adams_persistence_2017} and persistence-scale-space kernels~\cite{reininghaus_stable_2014}, as well as through filtration-based classification rules~\cite{kindelan_topological_2024, riihimaki_topological_2020}. To our knowledge, no prior work uses persistent homology as a prototype-selection criterion. We address this gap in this work.

\subsection{Contributions}
We present two complementary realizations of persistent homology as the selection criterion. \textit{Topological Prototype Selector} (TPS) uses sequential filtrations of a mixed point cloud of target and nontarget class observations as a first stage that captures between-class topological structure at the decision boundary, then refines the surviving points based on within-class topology. The \textit{Boundary-Conscious Topological Prototype Selector} (BoundaryTPS) encodes boundary proximity using vertex-weighted Rips filtration where points near the decision boundary enter the filtration before interior points. Both methods select prototypes by identifying persistence features at specified quantile lifetimes and recovering the data points that participate in those features.

Our main contributions are:
\begin{itemize}
    \item We introduce persistent homology as a prototype-selection criterion.
    \item We give two realizations, TPS and BoundaryTPS, that encode boundary information through different mechanisms --- mixed-class filtrations and vertex weights respectively --- and we show that the two mechanisms produce empirically different specializations.
    \item We benchmark both against seven classical baselines on fifteen real datasets across six downstream classifiers under nested cross-validation.
    \item We show empirically that the resulting prototype sets preserve $H_1$ persistence-diagram structure significantly better than every density-based or chained-decision baseline we test, and that TPS produces the most reproducible selection (highest mean fold-to-fold Jaccard) of any method evaluated.
    \item We characterize the trade-offs as both methods preserve the source set's class imbalance, empirically scale sub-quadratically in $n$, and reach competitive (though not leading) aggregate classification G-Mean, with non-trivial top-1 frequencies on (dataset, fold, classifier) combinations.
\end{itemize}

The remainder of this paper is organized as follows. Section~2 provides foundational background on the prototype selection problem, simplicial complexes, persistent homology, and weighted and multi-parameter persistence. Section~3 details the TPS and BoundaryTPS algorithms. Section~4 presents experimental evaluation on simulated and real-world tabular classification datasets. Section~5 closes with discussion of limitations and future directions.

\section{Foundational Concepts}
In this section, we provide some background on both the prototype selection problem and persistent homology. The concepts discussed are intended to be brief, high-level overviews. For more detailed discussion of the prototype selection problem we recommend~\cite{garcia_prototype_2012} and~\cite{olvera_lopez_prototype_2010}. For more detailed discussion of persistent homology and topological data analysis we recommend~\cite{edelsbrunner_computational_2010},~\cite{ghrist_barcodes_2008},~\cite{janes_theoretical_2025}, and~\cite{schenck_algebraic_2022}.

\subsection{Prototype Selection Problem}\label{sec:prototype_selection_problem}
Underlying all prototype selection methods is the dual problem of simultaneously seeking small prototype sets but retaining high downstream classification performance. Formally, the prototype selection problem is defined as
\begin{definition}\label{def:prototype_problem}
Given training data $\mathcal{X} = \{(x_1,y_1), \dots, (x_n,y_n)\}$ with class labels $y_i \in \mathcal{Y}= \{1,\dots,K\}$, the prototype selection problem is to find $\mathcal{S} \subset \mathcal{X}$ that minimizes $|\mathcal{S}|$ while maximizing classification performance of a downstream classifier trained on $\mathcal{S}$.
\end{definition}

Taxonomies for the methods are based on how they navigate this trade-off~\cite{garcia_prototype_2012}.

\textit{Condensation} methods such as condensed nearest neighbor (CNN) build $\mathcal{S}$ incrementally by appending misclassified instances until consistency is reached, at the cost of order-dependence~\cite{hart_condensed_1968}. \textit{Edition} methods such as edited nearest neighbor (ENN) do not as aggressively reduce the dataset, but act as noise edition methods by pruning the training set by discarding instances misclassified by their $k$ nearest neighbors~\cite{wilson_asymptotic_1972}. \textit{Hybrid} methods such as CNN+ENN chain the two strategies sequentially~\cite{devroye_probabilistic_1996}. The third algorithm in the family of decremental reduction optimization procedure algorithms (DROP3) pairs a noise filter with a decremental rule that retains an instance only when its removal harms the classification of its associates~\cite{wilson_reduction_2000}. Iterative case filtering (ICF) alternates ENN-style filtering with a reachability--coverage criterion that drops instances whose neighborhood is already covered~\cite{brighton_advances_2002}. \textit{Competence-based} methods such as third algorithm in the family of instance based learning algorithms (IB3) admit an instance only when its classification accuracy significantly exceeds the class prior, yielding implicit noise tolerance without tunable hyperparameters~\cite{aha_instance-based_1991}. \textit{Optimization-based} methods recast selection as a global problem, for example as a set cover~\cite{bien_prototype_2011}, class cover~\cite{manukyan_classification_2019, priebe_class_2003}, sparse subset selection on dissimilarity matrices~\cite{elhamifar_dissimilarity-based_2016}, or using greedy selection of prototypes using optimal transport (SPOTGreedy)~\cite{gurumoorthy_spot_2021}. \textit{Clustering-based} methods such as KMeans abandon selection entirely, instead generating synthetic class centroids rather than retain training instances~\cite{garcia_prototype_2012}.

The central principle shared among all prototype methods is the \textit{boundary-proximity principle}, which states that patterns near classification boundaries are considered to have higher classification contribution~\cite{olvera_lopez_prototype_2010, pekalska_prototype_2006}. The decision boundary itself can be viewed as a submanifold (or a union of submanifolds) separating classes, and that submanifold may have nontrivial structure at multiple spatial scales. Persistent homology is built to detect exactly this kind of structure across a range of scales, and to separate features that survive across scales from features that vanish under small perturbations. This capacity to characterize boundary geometry across scales motivates the use of persistent homology as the basis for prototype selection, as developed in Section~3.

\subsection{Simplicial Complexes}
\begin{figure}[tb]
\centering
\includegraphics[width=0.75\linewidth]{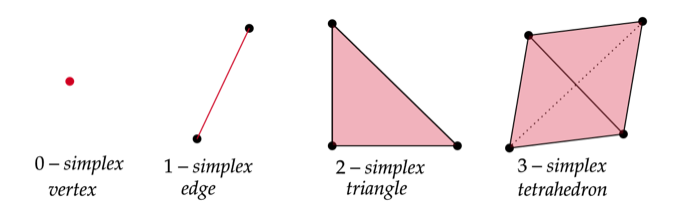}
\caption{Geometric representations of $q$-simplices. Each corresponds to their respective tetrahedral dimensional analogues.}
\label{fig:simplex_rep}
\end{figure}

Computational topology requires discrete representations called \textit{simplices} \cite{dey_computational_2022}. A \textit{simplex} is defined as follows 

\begin{definition}
A \textit{$q$-simplex}, $\sigma$, is the generalization of a tetrahedral region of space to $q$ dimensions. Given a $q$-simplex $\sigma$, a $d$-simplex $\tau$ with $0 \leq d \leq q$ and vertex set $|\mathcal{V}(\tau)| = d+1$ is called a \textit{$d$-face} of $\sigma$ denoted by $\tau \leq \sigma$. We also call $\sigma$ a \textit{$q$-coface} of $\tau$ denoted by $\sigma \geq \tau$. 
\end{definition}

Simplices provide discrete representations by encoding topological information through combinatorial structure. The vertex set $\mathcal{V}(\sigma)$ is the set of \textit{vertices} of $\sigma$ and the simplex $\sigma$ is said to be \textit{generated} by $\mathcal{V}(\sigma)$ \cite{schenck_algebraic_2022}. Geometrically, simplices correspond to either a vertex point (0-simplex), edges between vertices, (1-simplex), triangles (2-simplex), tetrahedra (3-simplex), and their higher-dimensional analogues as seen in Figure (\ref{fig:simplex_rep}).

In order to define homology groups of topological spaces, the notion of a \textit{simplicial complex} is central:

\begin{definition}
A simplicial complex $\mathcal{K}$ in $\mathbb{R}^n$ is a finite collection of simplices in $\mathbb{R}^n$ such that 
\begin{itemize}
\item [(i)] Given a $q$-simplex $\sigma \in \mathcal{K}$ and $\tau \leq \sigma$ then $\tau \in \mathcal{K}$, and
\item [(ii)] If $\sigma_1, \sigma_2 \in \mathcal{K}$ then $\sigma_1 \cap \sigma_2$ is either a face of both $\sigma_1$ and $\sigma_2$ or is empty. 
\end{itemize}
\end{definition}

\noindent Simplicial complexes provide the combinatorial structure on which homology and persistent homology are defined.

\subsection{Persistent Homology}
\textit{Homology} provides an algebraic framework for quantifying topological features using simplicial complexes. The \textit{homology group}, $H_k(\mathcal{X})$, is calculated by defining boundary maps between the \textit{chain} of different groups of formal linear combinations of $k$-simplices, finding the kernels and images of these maps, and then taking the quotient of the kernels by the images at each stage. For each dimension $k$, $H_k(\mathcal{X})$ characterizes the $k$-dimensional ``holes" in the topological space $\mathcal{X}$ \cite{carlsson_topology_2009, dey_computational_2022}. $H_0(\mathcal{X})$ measures connected components, $H_1(\mathcal{X})$ measures ``loops'' or 1-dimensional holes, $H_2(\mathcal{X})$ measures ``voids'' or 2-dimensional holes, and so on with $H_k(\mathcal{X})$ measuring $k$-dimensional analogues. 

The objective of persistent homology is to track how homology changes across scale values which vary incrementally, in a process known as a filtration. 

\begin{definition}
Let $\mathcal{K}$ be a simplicial complex. A \textit{filtration}, $\mathcal{F}$, on $\mathcal{K}$ is a succession of increasing \textit{sub-complexes} of $\mathcal{K}$, $\emptyset \subseteq \mathcal{K}_0 \subseteq \mathcal{K}_1 \subseteq \mathcal{K}_2\cdots\subseteq\mathcal{K}_n = \mathcal{K}$. In this case, $\mathcal{K}$ is called a filtered simplicial complex.
\end{definition}

\begin{figure}[tb]
\centering
\includegraphics[width=0.75\linewidth]{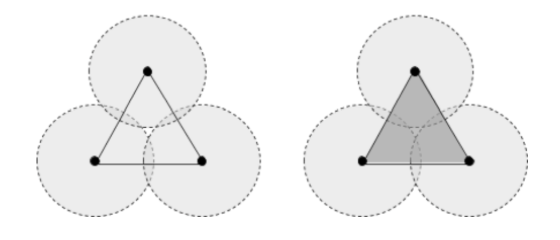}
\caption{Comparison of Čech complex (left) with Rips complex (right) at the same radius. The simplicial Čech complex consists of a hollow triangle since all three balls do not overlap, where the Rips complex is the solid triangle.}
\label{fig:cech_rips_compare}
\end{figure}

The filtration $\mathcal{F}$ on a filtered simplicial complex $\mathcal{K}$ is obtained by taking a collection of scaled values $\mathcal{E}_\mathcal{K}$ such that $0< \epsilon_0 < \epsilon_1 <\dots < \epsilon_n$, and the complex $\mathcal{K}_i$ corresponds to the value $\epsilon_i$. The set $\mathcal{E}_\mathcal{K}$ is called the \textit{filtration value collection} associated to $\mathcal{F}$. In general, scales along a filtrations correspond to metric covering balls with increasing radii that form covers of the set of data, $\mathcal{X}$.

Filtrations in TDA applications are dominated by two primary simplicial complex constructions, the \textit{Čech complex} and the \textit{Vietoris-Rips (Rips) complex} \cite{schenck_algebraic_2022}.

\begin{definition}
The \textit{Čech complex, $\check{C}_\epsilon(\mathcal{X})$}, is a simplicial complex with $q$-simplices if and only if there exists $\{x_i\}_{i =1}^{q+1} \in \mathcal{X}$ tuples of points such that $\bigcap_{i = 1}^{q+1} \overline{B_{\epsilon}(x_i)} \neq \emptyset$. That is, there exists a $q$-simplex if the intersections of the closed $\epsilon$-distanced balls centered at each of the $q+1$ points is non-empty. 
\end{definition}

Čech complexes act as a \textit{nerve complex} \cite{edelsbrunner_computational_2010}. Nerve complexes are abstract complexes that record the pattern of intersections between the points in $\mathcal{X}$. The \textit{nerve theorem} states that under conditions where Čech complexes form \textit{good covers}, the simplicial complex is homotopy equivalent to $\mathcal{X}$ \cite{schenck_algebraic_2022}. Rips complexes, in contrast, do not require all $\epsilon$-radius balls to intersect together to form a simplicial complex. 

\begin{definition}
The \textit{Vietoris-Rips complex, $R_\epsilon(\mathcal{X})$}, is a simplicial complex with $q$-simplices if and only if there exists $q+1$ sets of points of $\mathcal{X}$ such that $d(x_i, x_j) \leq \epsilon$ for all $x_i,x_j \in \mathcal{X}$ such that $i \neq j$ for distance metric $d(\cdot, \cdot)$. 
\end{definition}

The Rips complex only checks pairwise distances, and if every edge of a candidate simplex exists, the simplex is automatically included. This makes the Rips complex a \textit{flag complex}. If all edges exist, the higher-dimensional simplex is automatically included, making Rips complexes more computationally efficient to compute compared to Čech complexes. We provide a comparison of both simplicial complexes in Figure \ref{fig:cech_rips_compare}. 

In exchange for computational efficiency, Rips complexes provide only approximate topological accuracy to the underlying $\mathcal{X}$. The exact topological information is interleaved with the Vietoris--Rips construction through the
chain of inclusion maps $\check{C}_\epsilon(\mathcal{X}) \subseteq R_{2\epsilon}(\mathcal{X})
\subseteq \check{C}_{2\epsilon}(\mathcal{X})$~\cite{edelsbrunner_computational_2010, schenck_algebraic_2022}. Therefore, it is computationally feasible to determine the Rips complex at some comparable scale that captures the essential topological features of the Čech complex and vice-versa.

\begin{figure}[tb]
\centering
\includegraphics[width=\linewidth]{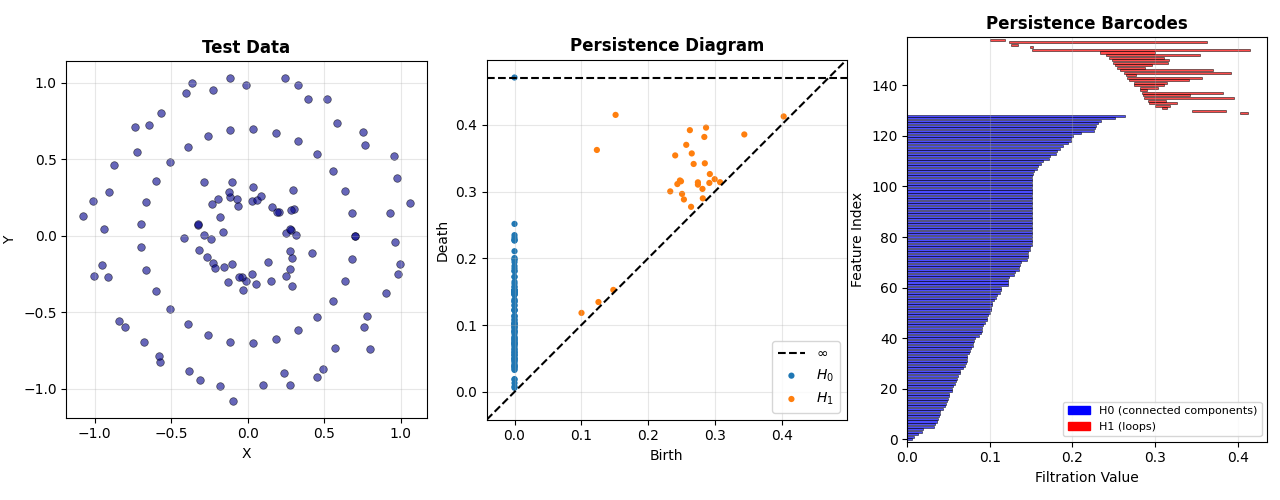}
\caption{Example of barcode and persistent homology visualizations. Left is the original data, middle is the corresponding persistent homology diagram (birth, death) pairs plotted, and right is the barcodes for specific filtration values.}
\label{fig:ph_barcode_example}
\end{figure}

Given $\{\mathcal{K}_i\}_{i=0}^n$, the inclusion maps $\mathcal{K}_i \hookrightarrow \mathcal{K}_j$ induce homomorphisms $H_k(\mathcal{K}_i) \rightarrow H_k(\mathcal{K}_{i'})$ on homology groups. Essentially, each level change from $i$ to $i'$ has the potential to correspond to either a ``birth'' or a ``death'' of a topological feature. This yields a sequence of vector spaces connected by linear maps called a \textit{persistence module}. Persistence modules decompose into ``birth'' and ``death'' pairs, where the length of the interval of a (birth, death) pair, is called its \textit{persistence} or \textit{lifetime} \cite{schenck_algebraic_2022, carlsson_topology_2009, kindelan_topological_2024}. Long persistence features track topological features in the data, while short persistence corresponds to noise. Persistence can be graphically represented through \textit{barcodes} or by plotting (birth, death) pairs on \textit{persistence diagrams} \cite{ghrist_barcodes_2008, carlsson_topology_2009}. We display both styles of visualizations on a worked example in Figure (\ref{fig:ph_barcode_example}).

\subsubsection{Weighted Persistent Homology}
Not every point in a point cloud is equally informative \cite{meng_weighted_2020, anand_weighted_2020}. Points near a classification boundary may carry more discriminative information than points deep in a class interior, or points in high-density regions may be more representative of an underlying
distribution than isolated outliers. Standard persistent homology cannot distinguish between such points as every point participates in the filtration from its onset. 

\textit{Weighted persistent homology} fixes this by attaching points with a non-negative weight that delays when it enters the filtration \cite{ren_weighted_2018}. Points assigned low weight enter early, while high-weight points are delayed and enter only at coarser scales. The weight vector thus provides a principled mechanism for
encoding auxiliary information (such as function values, density
estimates, or domain-specific importance scores) directly into the
topological computation, without changing the underlying point cloud geometry. 

\begin{definition}\label{def:weighted_rips}
Let $\mathcal{X} = \{x_1, \dots, x_n\}$ be a finite point set with
distance function $d(\cdot,\cdot)$, and let
$\mathbf{w} = (w_1, \dots, w_n)$ be a vector of non-negative vertex
weights. The \textit{vertex-weighted Rips filtration}
$R^\mathbf{w}_\epsilon(\mathcal{X})$ is defined by:
\begin{itemize}
    \item[(i)] Vertex $x_i$ enters at scale $w_i$, i.e.,
        $x_i \in R^\mathbf{w}_\epsilon(\mathcal{X})$ if and only if
        $\epsilon \geq w_i$.
    \item[(ii)] An edge $\{x_i, x_j\}$ enters at scale
        $\max\bigl(d(x_i, x_j), w_i,w_j\bigr)$.
    \item[(iii)] A higher simplex $\sigma$ enters at the maximum
        entrance value among all its edges.
\end{itemize}
\end{definition}

Condition~(ii) ensures that an edge cannot appear before both of its
endpoints exist in the complex. When $w_i = 0$ for all~$i$, the
construction reduces to the standard Rips filtration. Computationally, the vertex-weighted Rips filtration is equivalent to
computing standard Rips persistence on the modified distance matrix
\begin{equation}\label{eq:modified_distance}
    \tilde{D}[i,j] =
    \begin{cases}
        w_i & \text{if } i = j, \\
        \max\bigl(d(x_i, x_j),\; w_i,\; w_j\bigr) & \text{if } i \neq j,
    \end{cases}
\end{equation}
since the diagonal entries encode vertex birth times and the off-diagonal
entries encode edge entrance times \cite{ren_computational_2021}.

\subsection{Multi-Parameter Persistence and Sequential Filtration}
Many applications require simultaneous consideration of filtrations across multiple scale parameters \cite{xia_multidimensional_2015, vipond_multiparameter_2021}. A \textit{bifiltration} extends the single-parameter filtration framework by indexing sub-complexes $\mathcal{K}_{i,j}$ along two filtration parameters, where increasing either parameter yields a larger sub-complex with inclusion maps commuting \cite{janes_theoretical_2025}. The formal definition is as follows:

\begin{definition}
    A bifiltration of a topological space $X$ is a family of subsets $\{\mathcal{K}_{i,j}\}_{i,j \in \mathcal{E}_{i,j}}$ such that the subsets satisfy the following:
    \begin{itemize}
        \item For any fixed $i,j \mapsto \mathcal{K}_{i,j}$ is a filtration (increasing with $i$), and
        \item For any fixed $j,i \mapsto \mathcal{K}_{i,j}$ is also a filtration (increasing with $j$).
    \end{itemize}
\end{definition}

Figure (\ref{fig:commute_diagram}) illustrates this structure. Each $\mathcal{K}_{i,j}$ is nested within both $\mathcal{K}_{i+1,j}$
and $\mathcal{K}_{i,j+1}$, forming a grid of sub-complexes connected by compatible inclusion maps.

\begin{figure}[tbhp]
\centering
\begin{tikzcd}[row sep=large, column sep=large]
\emptyset \arrow[r] & \mathcal{K}_{0,0} \arrow[r] \arrow[d] & \mathcal{K}_{1,0} \arrow[r] \arrow[d] & \mathcal{K}_{2,0} \arrow[d] \\
& \mathcal{K}_{0,1} \arrow[r]   & \mathcal{K}_{1,1} \arrow[r]   & \mathcal{K}_{2,1} \arrow[r] & \mathcal{K} \\
\end{tikzcd}
\caption{An example of a commutative diagram for a bifiltration of $\mathcal{K}$ across two parameters. Each $\mathcal{K}_{i,j}$ represents a sub-complex.}
\label{fig:commute_diagram}
\end{figure}

Many of the fundamental theorems for filtrations do not have analogous variants for bifiltrations. Unlike single-parameter persistence, multi-parameter persistence modules do not admit a complete discrete invariant analogous to barcodes \cite{schenck_algebraic_2022, scaramuccia_computing_2020}. The classification of finitely generated modules over multivariable polynomial rings is substantially more complex than the single-variable case, and no direct analogue of the persistence diagram exists for the full bifiltration \cite{schenck_algebraic_2022}. One approach to recovering tractable invariants is to \textit{slice} the bifiltration along one parameter at a fixed level of the other, reducing each slice to a standard single-parameter persistence computation with a well-defined barcode.

\section{Topological Prototype Selection}

Let $\mathcal{X} = \{(x_i, y_i)\}_{i=1}^n \subset \mathbb{R}^d$ be training data with class labels $y_i \in \mathcal{Y} = \{1, \dots, C\}$. For target class $c$, let $\mathcal{X}_c = \{x_i \mid y_i = c\}$ denote the target-class subset and $\mathcal{X}_{\neg c} = \{x_i \mid y_i \neq c\}$ the complement. We present two realizations of topological prototype selection. TPS encodes boundary information implicitly by constructing a filtration on a mixed-class point cloud and then further refinement, while BoundaryTPS encodes it explicitly through vertex weights that delay interior points relative to boundary-proximate ones in a single-class filtration.
 
All relevant simplicial complexes were built and homology calculated using the GUDHI library with maximum edge length set to the maximum pairwise distance and maximum simplicial dimension set to $p + 1$, where $p$ is the homology dimension \cite{maria_rips_2025, tinarrage_weighted_2025}. We begin first with our boundary-conscious topological prototype selector.

\subsection{Boundary-Conscious Topological Prototype Selection (BoundaryTPS)}

The intuition for BoundaryTPS is that proximity to the decision boundary should be encoded directly into when a point participates in the topological computation. Every target-class point becomes a candidate prototype, and each is assigned a weight that summarizes its distance to the nearest non-target points. Ideally small weights are given for points sitting near the decision boundary to enter the filtration early, with large weights for points deep in the class interior delaying entry. The weighted Rips filtration then resolves boundary-region topology first, since edges between two low-weight points enter the filtration at small filtration values, while edges incident to high-weight interior points are delayed until the filtration parameter grows to admit them. Persistence features born from boundary-region structure therefore appear at small lifetimes, and lifetime-quantile selection cleanly recovers the boundary-proximate vertex set. 

Boundary-conscious topological prototype selector (BoundaryTPS) leverages a single-parameter weighted filtration introduced in Definition \ref{def:weighted_rips} to encode boundary proximity information. For each target-class point $x_i \in \mathcal{X}_c$, the instance weight is computed as the sum of distances to its $k$ nearest neighbors (kNN) in the nontarget classes. That is, 
\begin{equation}\label{eq:boundary_weight}
    w_i = \sum_{j \in \mathrm{kNN}_{\neg c}(x_i)} d(x_i, x_j)
\end{equation}
where $\mathrm{kNN}_{\neg c}(x_i)$ denotes the indices of the $k$ nearest nontarget neighbors of $x_i$ under distance metric $d$. Points near the decision boundary receive small weights, while interior points far from any nontarget observation receive large weights.

The modified distance matrix
\begin{equation}\label{eq:btps_modified_distance}
    \tilde{D}_{ij} = \begin{cases}
        w_i & \text{if } i = j, \\
        \max\bigl(d(x_i, x_j),\; w_i,\; w_j\bigr) & \text{if } i \neq j
    \end{cases}
\end{equation}
is constructed over $\mathcal{X}_c$ and a Vietoris-Rips filtration is built on $\tilde{D}$. In this filtration, an edge $(x_i, x_j)$ cannot appear until the filtration parameter $\epsilon$ exceeds $\max(d(x_i, x_j), w_i, w_j)$. Edges between two low-weight boundary points potentially enter the filtration governed primarily by their actual pairwise distance, since $d(x_i, x_j) \geq \max(w_i, w_j)$ is common when both points are near the decision boundary. Edges involving high-weight interior points are delayed until $\epsilon$ reaches the larger weight, regardless of the actual pairwise distance between the points. The effect is that boundary-region topology is resolved first in the filtration, while interior structure appears only at coarser scales.

Persistent homology in dimension $p$ is computed on the weighted filtration, yielding a set of persistence features $F = \{(b_i, d_i, \ell_i)\}$ where $\ell_i = d_i - b_i$ is the lifetime of feature $i$ \cite{maria_persistent_2025}. Let the collection of all persistence lifetimes be denoted $L = \{\ell_i\}_{i=1}^m$. Given a user-specified quantile $q \in (0,1)$, the persistence feature whose lifetime is closest to $\ell_q$ is selected. We calculate the quantile-lifetime through function $\mathrm{QuantInt}$ defined as,
\begin{equation} \label{eq:neighbor_quantile}
\ell_q = \mathrm{QuantInt}(L) = \begin{cases}
\ell_{(1)} & \text{if } q = 0 \\
\ell_{(n)} & \text{if } q = 1 \\
(1-\gamma) \cdot \ell_{(z)} + \gamma \cdot \ell_{(z+1)} & \text{otherwise}
\end{cases}
\end{equation} where $\ell_{(z)} $ represents the $z^{th}$ ordered lifetime, and \begin{align}
h &= (n-1) \cdot q + 1 \\
z &= \lfloor h \rfloor \\
\gamma &= h - z \label{eq:gamma}
\end{align} where Equations (\ref{eq:neighbor_quantile})--(\ref{eq:gamma}) identify the $q^{th}$ quantile persistence lifetime of the distribution of all lifetime values.  

The parameter $q$ acts as a geometric regularization parameter that controls how aggressively BoundaryTPS prunes. Lower quantiles select shorter-lived features which in turn correspond to prototype sets with observations closer to the boundary regions. Larger quantiles will retain features associated with broader topological structure further from the boundary as more observations participate in the filtration. A tolerance parameter $\tau$ is introduced to provide additional control over the size of the selected feature set and may improve robustness by reducing sensitivity to the exact configuration of the persistence diagram. When the tolerance parameter $\tau = 0$, the exact feature(s) whose lifetime is closest to $\ell_q$ are selected (with ties included). When $\tau > 0$, all features with lifetime in the window $[\ell_q(1 - \tau),\; \ell_q(1 + \tau)]$ are selected, which softens the discrete nature of feature selection by including multiple features near the target lifetime. If the tolerance window captures no other features, the algorithm falls back to the single closest match.

Once features are selected, we identify  which data points participate in those features. For $H_0$ (connected components), a Union-Find data structure with full member tracking replays the filtration edges in order up to the maximum death value among the selected features which yields exact component membership at each merge event \cite{glisse_fast_2023}. Each time two connected components merge all vertices in both components are recorded. The vertices that participated in any merge event constitute the prototype set for the target class. For higher homology dimensions ($p \geq 1$), the algorithm instead collects all vertices of $(p+1)$-simplices (e.g., triangles for $H_1$) that enter the filtration up to the maximum death value of the selected features. Since only target-class points are present in the filtration, no post-filtering is required. Complete pseudocode is provided in the Appendix.

\subsection{Topological Prototype Selection (TPS)}

The intuition for TPS is that boundary information is already present in the geometry between the target class and its nearest non-target neighbors. Essentially, no weighting is needed if those non-target points are explicitly included in the point cloud. Stage~1 builds that mixed point cloud and runs a standard Rips filtration on it, so the persistence features capture between-class structure at the local decision boundary. Selected features are pulled back to their target-class vertices, discarding the non-target scaffolding. A second Rips filtration is built on the survivors, with feature selection at the mean lifetime, retains an interior-typical sample that supplements the boundary-relevant points from Stage~1. The two stages together produce a prototype set with both boundary coverage and interior support.

TPS takes an alternative approach to incorporating boundary information. Rather than encoding boundary proximity through vertex weights on a single-class filtration, TPS constructs a mixed point cloud containing both target and nontarget observations and applies two sequential Vietoris-Rips filtrations, motivated by the slicing of bifiltrations described in Section 2.4. The first filtration captures between-class topological information, and the second captures within-class information of the surviving points.

\subsubsection{Stage 1: Boundary-Neighborhood Filtration}

For each target class $c$, TPS constructs a mixed point cloud $B_c$ consisting of all target-class points $\mathcal{X}_c$ together with their $k$ nearest nontarget neighbors:
\begin{equation}\label{eq:mixed_cloud}
    B_c = \mathcal{X}_c \cup \bigcup_{x_i \in \mathcal{X}_c} \mathrm{kNN}_{\neg c}(x_i)
\end{equation}
where the union over nontarget neighbors is taken over unique indices to avoid duplication. The inclusion of nontarget points in the point cloud means that the resulting filtration directly encodes the geometry of the between-class boundary region. Restricting to $k$ nearest nontarget neighbors rather than all of $\mathcal{X}_{\neg c}$ focuses the filtration on the local decision boundary, where between-class topological structure is most informative for prototype selection. As $k$ increases, nontarget points further from the boundary are included, potentially diluting the boundary signal.

A standard Vietoris-Rips filtration is built on the pairwise distance matrix of $B_c$, and persistent homology in dimension $p$ is computed. Feature selection uses $\mathrm{QuantInt}$ (Equation \ref{eq:neighbor_quantile}) with the quantile parameter $q$ but without the tolerance window (i.e., selecting only the feature(s) whose lifetime is closest to the target quantile, with ties included).

Vertex participation is resolved using the same mechanisms as BoundaryTPS (Union-Find replay for $H_0$, simplex enumeration for $p \geq 1$). After vertex resolution, the surviving set is reduced to retain only target-class points, discarding the nontarget neighbors that were included to define the boundary geometry.

\subsubsection{Stage 2: Interior Filtration}

Stage 2 takes the target-class survivors from Stage 1 and builds a new Vietoris-Rips filtration on their pairwise distance matrix. This second filtration captures the within-class topological structure of the boundary-relevant points identified in Stage 1.

Feature selection in Stage 2 uses the mean lifetime rather than a quantile:
\begin{equation}\label{eq:avgint}
    \ell_{\mathrm{mean}} = \frac{1}{|F'|} \sum_{f_i \in F'} \ell_i
\end{equation}
where $F'$ is the set of persistence features from the Stage 2 filtration. The feature(s) whose lifetime is closest to $\ell_{\mathrm{mean}}$ are selected, with ties included. Selecting at the mean targets features of typical topological scale within the class, avoiding both short-lived noise and the longest-lived global structure that may not correspond to boundary-relevant geometry.

Vertex participation resolution proceeds identically to Stage 1, yielding the final prototype set for target class $c$. Complete pseudocode is provided in the Appendix.

\subsection{Algorithmic Comparisons and Remarks}

TPS and BoundaryTPS share two implementation details that we describe here for reproducibility. Table~\ref{tab:method_comparison} summarizes the structural differences. Neither design dominates the other empirically, and relative performance depends on the downstream classifier, the dataset geometry, and which side of the boundary/interior trade-off is preferred for a given application.

Features that never die in the filtration (essential features) have their death value set to the maximum pairwise distance in the relevant distance matrix (e.g. $\max(D)$ or $\max(\tilde{D})$). This truncation ensures that all features have finite lifetimes and can participate in the quantile computation and mean computation. When the number of nontarget points is smaller than the requested neighborhood size $k$, the actual neighborhood size is clamped at $k_{\mathrm{actual}} = \min(k, |\mathcal{X}_{\neg c}|)$.

In the event where multiple features share the same lifetime as the closest match to the target quantile or mean, all tied features are included. This inclusive tie-breaking ensures that topologically equivalent features are not arbitrarily excluded.

For $H_0$ persistence, each feature is a merge of two connected components at a specific filtration value. Both methods replay the filtration edges in ascending order of their entrance values using a Union-Find data structure with path compression and union by rank. Each component root maintains the complete set of original dataset indices of its members. When two components merge and the merge corresponds to a selected feature (i.e., the edge entrance value does not exceed the maximum death value among selected features), all member vertex indices of both components are recorded. The union over all such merges is the participating vertex set.

\begin{remark}
For $H_0$, the Union--Find replay returns exactly the set of vertices that participate in connected-component merges with death time at most $d_{\max} = \max\{d_f : f \in F_{\mathrm{sel}}\}$. Each merge event records the full membership of both components at the moment of merge, and the union over selected merges is the prototype set. No approximation is introduced at this step.
\end{remark}

For higher homology dimensions ($p \geq 1$), vertex participation is resolved differently. The algorithm enumerates all $(p+1)$-simplices from the simplex tree that enter the filtration up to the maximum death value of the selected features, and collects all vertices of those simplices. We note that for $p \geq 1$, the algorithm returns the vertex set of every $(p+1)$-simplex admitted by the filtration up to $d_{\max}$, not the representative cycle of any specific selected feature. Exact cycle recovery would require either (i) reading off representatives from the persistence pairing (computationally expensive and basis-dependent) or (ii) optimizing for minimal homologous cycles, which is NP-hard in general~\cite{chen_hardness_2011}. 

The simplex-enumeration shortcut used here is a practical proxy. It retains the support region in which the selected $H_p$ features live, at the cost of returning additional points that lie within that support but do not participate in any single representative. The composition ablation in Section~\ref{section:ablation} characterizes the resulting prototype set, and the topology-preservation evaluation in Section~\ref{sec:topology_preservation} measures whether this proxy is sufficient for downstream persistence-diagram reconstruction.

\begin{table}[tb]
\centering
\caption{Structural comparison of TPS and BoundaryTPS.}
\label{tab:method_comparison}
\small
\begin{tabular}{lll}
\hline
\hline
\textbf{Aspect} & \textbf{TPS} & \textbf{BoundaryTPS} \\
\hline
Stages & Two & One \\
Point cloud & Mixed (target + nontarget) & Target only \\
Boundary encoding & Implicit (mixed classes) & Explicit (vertex weights) \\
Distance matrix & Standard $D$ & Weighted $\tilde{D}$ \\
Feature selection & Quantile (Stage 1), Mean (Stage 2) & Quantile with tolerance $\tau$ \\
Hyperparameters & $k$, $q$, $p$ & $k$, $q$, $\tau$, $p$ \\
Post-filtering & Intersect with target indices & Not needed \\
\hline
\hline
\end{tabular}
\end{table}

At this point, we offer a stability \emph{intuition} rather than a uniform theorem.

\begin{remark}
Let $\mathcal{X}_c$ be the target-class point set and $\mathcal{X}_c'$ a perturbation with Hausdorff distance $d_H(\mathcal{X}_c, \mathcal{X}_c') \le \delta$. The classical
persistence-diagram stability theorem gives
\[
d_B\!\left(\mathrm{Dgm}(R(\mathcal{X}_c)),\,\mathrm{Dgm}(R(\mathcal{X}_c'))\right) \le \delta
\]
in bottleneck distance~\cite{cohen-steiner_stability_2007}, which should apply directly to the unweighted Vietoris--Rips constructions used by TPS.  
\end{remark}

For BoundaryTPS's weighted filtration, an analogous diagram-level stability property holds whenever the vertex weights vary continuously in the point coordinates. The $k$-nearest-non-target sum satisfies this away from configurations at which nearest-neighbor identities tie. At such ties the neighbor identities (and therefore the weight values) can change
discontinuously. 

Quantile-based feature selection inherits a parallel caveat. The selected lifetime
is continuous in the diagram between discrete events, but the selected \emph{set} of features can jump when features cross the quantile threshold or when persistence-pairing ties are broken. 

Under perturbations that avoid these discrete events, the persistence diagrams, selected lifetimes, and selected feature sets vary continuously. The empirical fold-to-fold Jaccard results in the following Section are consistent with this. The vertex-recovery step is itself set-valued and discrete, contributing additional fold-to-fold variation that is not controlled by the diagram-level bound.

We close this section with a conjecture on the time complexity. 

\begin{conjecture}[Time complexity]
For a target class of size $n_c$ with $n_{\neg c}$ non-target points in dimension $d$, $H_0$ persistence computation for BoundaryTPS is $\mathcal{O}(n_c^2 d + n_c k \log n_{\neg c})$ in the worst case. $\mathcal{O}(n_c^2 d)$ for the modified pairwise distance matrix and $\mathcal{O}(n_c k \log n_{\neg c})$ for the $k$-nearest-non-target lookups. TPS adds Stage~1 over a mixed cloud of size $|B_c| \leq n_c + n_c k$ and Stage~2 over the surviving target subset, both bounded by the same expression with $n_c$ replaced by the larger constant. Higher-homology computation in either method is bounded by the cost of enumerating $(p+1)$-simplices admitted up to the truncation, which is at worst $\mathcal{O}(n_c^{p+2})$ but in practice substantially smaller due to GUDHI's pruning of edges above the filtration threshold.
\end{conjecture}

\section{Results}
\begin{table}[tb]
\centering
\small
\caption{Prototype selection methods and hyperparameter search grids used in the inner cross-validation loop. $\lfloor\sqrt{n}\rfloor$ denotes the integer square root of the training set size. Methods marked with $\dagger$ generate synthetic prototypes rather than selecting original training instances.}
\label{tab:method_grids}
\resizebox{\textwidth}{!}{%
\begin{tabular}{lllll}
\hline\hline
\textbf{Method} & \textbf{Type} & \textbf{Parameter} & \textbf{Description} & \textbf{Search Grid} \\
\hline
\multirow{3}{*}{TPS} & \multirow{3}{*}{Topological}
 & $k$ & Nontarget neighbors & $\{1, 3, 5, 10, 15, \lfloor\sqrt{n}\rfloor\}$ \\
 & & $q$ & Lifetime quantile & $\{0.025, 0.05, 0.10, 0.12, 0.15, 0.20, 0.25, 0.30, 0.35, 0.40, 0.50\}$ \\
\hline
\multirow{4}{*}{BoundaryTPS} & \multirow{4}{*}{Topological}
 & $k$ & Nontarget neighbors & $\{1, 3, 5, 10, 15, \lfloor\sqrt{n}\rfloor\}$ \\
 & & $q$ & Lifetime quantile & $\{0.025, 0.05, 0.10, 0.12, 0.15, 0.20, 0.25, 0.30, 0.35, 0.40, 0.50\}$ \\
 & & $\tau$ & Tolerance window & $\{0, 0.01, 0.025, 0.05, 0.10\}$ \\
\hline
CNN \cite{hart_condensed_1968} & Condensation & $k$ & Neighborhood size & $\{1, 3, 5, 7, 10, 15, \lfloor\sqrt{n}\rfloor\}$ \\
\hline
CNN+ENN \cite{devroye_probabilistic_1996} & Hybrid & $k_{\mathrm{ENN}}$ & ENN neighborhood size & $\{1, 3, 5, 7, 10, 15, \lfloor\sqrt{n}\rfloor\}$ \\
\hline
DROP3 \cite{wilson_reduction_2000} & Hybrid & $k$ & Neighborhood size & $\{1, 3, 5, 7, 10, 15, \lfloor\sqrt{n}\rfloor\}$ \\
\hline
ICF \cite{brighton_advances_2002}& Hybrid & $k$ & Neighborhood size & $\{1, 3, 5, 7, 10, 15, \lfloor\sqrt{n}\rfloor\}$ \\
\hline
IB3 \cite{aha_instance-based_1991} & Competence & $\lambda$ & Thresholds & $\lambda_{\text{accept}} = 0.9$ \\ &&&& $\lambda_{\text{reject}} = 0.7$ \\
\hline
KMeans \cite{garcia_prototype_2012}$^\dagger$ & Clustering & $m$ & Prototypes per class & $\{10, 20, 50\}$ \\
\hline
SPOTGreedy \cite{gurumoorthy_spot_2021}$^\dagger$ & Optimization & $m$ & Prototypes per class & $\{10, 20, 50\}$ \\
\hline\hline
\end{tabular}
}
\end{table}

We evaluate the performance of TPS across various classification settings. The evaluation is organized around five questions, in order of priority for the contribution: (i) do the proposed methods preserve the topological structure of the training set better than existing methods (Section~\ref{sec:topology_preservation}); (ii) are their selections stable across fold perturbation (Section~\ref{sec:selection_stability}); (iii) what classification performance do their prototype sets support (Section~\ref{sec:classification_performance}); (iv) how do they treat class imbalance (Section~\ref{sec:class_imbalance_behavior}); and (v) what do they cost (Section~\ref{sec:computational_cost})? Questions (i) and (ii) are the criteria most distinctive of a topology-based selector, and we evaluate them first. Question (iii) is the criterion the existing literature most often optimizes; we report it without expecting topology-based selection to dominate it.

We opt for a nested cross-validation (CV) framework across all experiments. A stratified 10-fold outer loop estimates generalization performance, while a stratified 5-fold inner loop selects hyperparameters for each method. Table~\ref{tab:method_grids} lists each prototype selection method and its hyperparameter search grid. The same nested protocol applies to every method. We report G-Mean rather than accuracy as the primary metric because its multiplicative form penalizes any configuration that misclassifies an entire class \cite{japkowicz_assessment_2013, bekkar_evaluation_2013}. This choice is particularly important for prototype selection, where reduction-aggressive configurations risk disproportionately eliminating minority class instances \cite{garcia_proposal_2006}. 
 
Tuning hyperparameters for prototype methods requires balancing G-Mean against data reduction, as optimizing either criterion in isolation produces degenerate solutions. Pure G-Mean optimization can select configurations that retain nearly the entire training set \cite{kuncheva_fitness_1997}, while tuning for pure reduction can discard discriminative instances. Recent literature on evolution-based instance selection has used linear scalarization fitness functions such as $\alpha \cdot \mathrm{G\text{-}Mean} + (1-\alpha) \cdot \mathrm{Reduction}$ \cite{cano_using_2003}, but the limitations of single-value scalarizations are well-known \cite{das_closer_1997, luo_pareto-optimal_2025}. 

We instead use a Pareto-based optimization to select optimal hyperparameters. For each candidate hyperparameter configuration evaluated in the inner CV loop, we record the minimum G-Mean and minimum reduction rate across all inner folds. We then identify the \textit{Pareto-optimal} configurations by applying rank normalization to map each objective to $[0, 1]$ independent of scale, and select the configuration closest in distance to the ideal point (rank 0 on both objectives). This closest-to-ideal criterion selects the most balanced compromise on the Pareto front and is grounded in compromise programming and the TOPSIS decision framework \cite{zeleny_compromise_1973, yu_class_1973, hwang_multiple_1981}. Rank normalization, rather than min-max normalization, ensures robustness to outlier configurations and produces a uniform objective-space distribution regardless of the differences in raw value ranges \cite{conover_rank_1981}.
 
1-nearest neighbor (1-NN), 3-nearest neighbor (3-NN), 5-nearest neighbor (5-NN), linear-kernel support vector machine (LinearSVM), random forest (RF), and a single-hidden-layer multilayer perceptron (MLP) are trained on each prototype set to evaluate generalization across classifier families. Classifier hyperparameters are held fixed across all experiments, and only prototype selector hyperparameters are tuned via the inner loop to assess the quality of the selected prototype set rather than jointly optimize selection and classification. While TPS can be performed over any homology group, we focus exclusively on $H_0(\mathcal{X})$ unless otherwise specified. 
 
\subsection{Performance on Simulated Datasets}
\label{sec:simulated}

\begin{figure}[b]
    \centering
    \includegraphics[width=\linewidth]{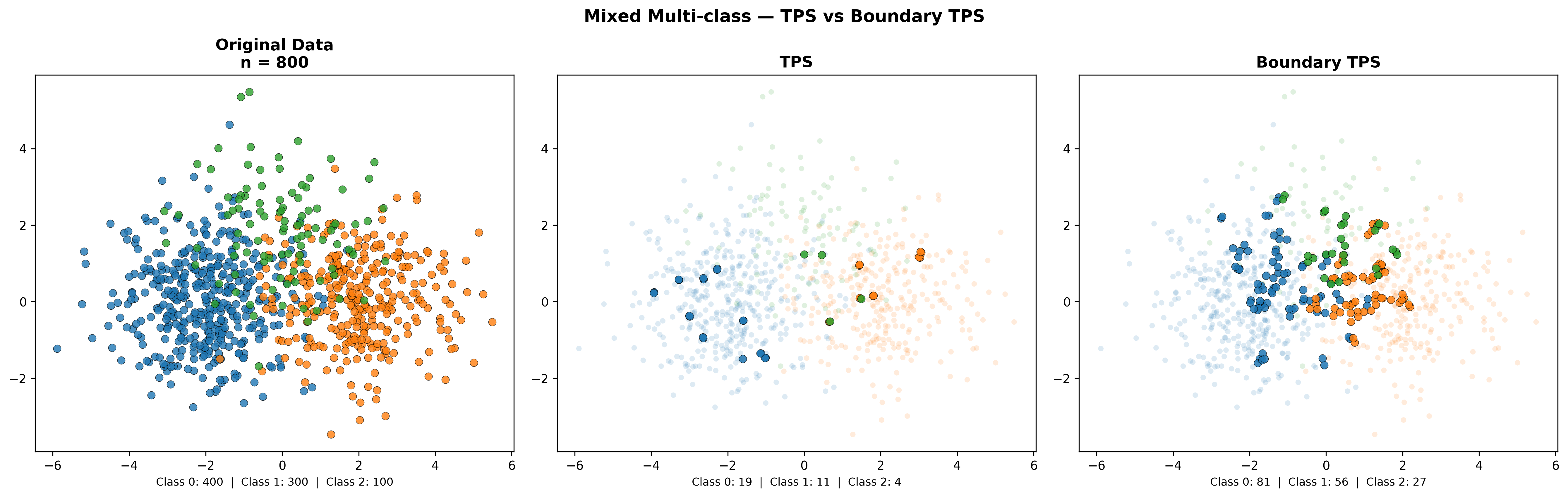}
\caption{A single-fold prototype selection on Mixed Multiclass (3-class, class distribution 400/300/100, IR $= 4.0$) for a representative fold. \emph{Left}: the original training set. \emph{Center}: TPS prototypes ($n = 72$; class counts 42/21/9). \emph{Right}: BoundaryTPS prototypes ($n = 164$; class counts 81/56/27).}
\label{fig:mixed_proto_comparison}
\end{figure}

We first evaluate TPS and BoundaryTPS on nine simulated datasets exploring between-class separation (well-separated to heavily overlapping clusters), noise level (moderate to high perturbation of nonlinear decision boundaries), and class balance (balanced to 4:1 imbalance ratio) as a performance check to better understand performance differences between the two methods. Datasets were generated in $\mathbb{R}^2$ for visualizations. Representative visualizations for a single fold are available in the Appendix. Table~\ref{tab:dataset_description} summarizes the characteristics of each dataset and the baseline G-Mean achieved by training on the full dataset with no prototype selection applied.
 
\begin{table}[tb]
\centering
\caption{Descriptions of the nine simulated datasets. Baseline G-Mean is the mean across 10 folds when training on the full dataset with no prototype selection. The imbalance ratio (IR) for each dataset is also shown.}
\label{tab:dataset_description}
\small
\resizebox{\textwidth}{!}{%
\begin{tabular}{lccccccccccc}
\hline\hline
\textbf{Dataset} & \textbf{n} & \textbf{Classes} & \textbf{Class Ratio} & \textbf{IR} & \multicolumn{6}{c}{\textbf{Baseline G-Mean}} \\
\cline{6-11}
 & & & & & \textbf{1-NN} & \textbf{3-NN} & \textbf{5-NN} & \textbf{SVM} & \textbf{RF} & \textbf{MLP} \\
\hline
Well-Sep.\ Blobs        & 600 & 3 & 200:200:200     & 1.00 & 1.000 & 1.000 & 1.000 & 1.000 & 1.000 & 1.000 \\
Overlapping Blobs       & 800 & 4 & 200:200:200:200 & 1.00 & 0.634 & 0.677 & 0.706 & 0.728 & 0.699 & 0.731 \\
Overlapping Clusters    & 600 & 3 & 200:200:200     & 1.00 & 0.475 & 0.473 & 0.498 & 0.540 & 0.456 & 0.527 \\
Two Moons (Mod.\ Noise) & 500 & 2 & 250:250         & 1.00 & 0.996 & 0.994 & 0.996 & 0.858 & 0.988 & 0.980 \\
Two Moons (High Noise)  & 500 & 2 & 250:250         & 1.00 & 0.817 & 0.855 & 0.869 & 0.836 & 0.867 & 0.869 \\
Circles (Mod.\ Noise)   & 500 & 2 & 250:250         & 1.00 & 0.982 & 0.984 & 0.984 & 0.484 & 0.970 & 0.984 \\
Circles (High Noise)    & 500 & 2 & 250:250         & 1.00 & 0.736 & 0.782 & 0.787 & 0.517 & 0.776 & 0.818 \\
Imbalanced (80/20)      & 500 & 2 & 400:100         & 4.00 & 0.705 & 0.703 & 0.721 & 0.805 & 0.734 & 0.778 \\
Mixed Multiclass        & 800 & 3 & 400:300:100     & 4.00 & 0.664 & 0.691 & 0.693 & 0.750 & 0.695 & 0.759 \\
\hline\hline
\end{tabular}
}
\end{table}

Table~\ref{tab:simulated_results} in the Appendix reports the aggregated G-Mean $\pm$ standard deviation across the evaluation folds. Across the nine simulated datasets, both methods achieve substantial dataset reductions while typically sustaining classification performance close to the full-data baseline. Reduction rates fall between 81.7\% and 93.7\% for TPS and between 78.6\% and 96.7\% for BoundaryTPS, with prototype selection completing in 0.03 to 0.11 seconds for TPS and 0.02 to 0.03 seconds for BoundaryTPS across all datasets. 

The differences track the structural distinction between the two methods. 

TPS's second stage adds interior observations that preserve within-class topology. BoundaryTPS's vertex-weighted single-stage filtration concentrates prototypes near cross-class density. Which is better depends if broad interior support or a sharp boundary resolution is needed from the classifier on a given dataset. The Imbalanced (80/20) dataset is the clearest case. Every classifier improves over its baseline under TPS, with $\Delta$G-Mean ranging from $+1.5$ (MLP) to $+12.4$ (1-NN), while BoundaryTPS is mixed, with moderate gains for 3-NN ($+0.4$) and 5-NN ($+2.5$) offset by substantial drops for Random Forest ($-11.2$) and MLP ($-7.9$). Mixed Multiclass shows a similar pattern as TPS produces smaller-magnitude deltas across classifiers ($-3.4$ to $+2.3$) than BoundaryTPS, which degrades on every classifier ($-11.1$ to $-1.8$). On the more uniformly overlapping cases, Overlapping Blobs and Overlapping Clusters, both methods perform comparably, with most deltas modest in magnitude (TPS: $-5.9$ to $+0.5$; BoundaryTPS: $-7.7$ to $+3.0$). We highlight specifically the selection of both methods for this dataset in Figure~\ref{fig:mixed_proto_comparison}, where TPS retains 72 points across the three classes (42/21/9; 91.0\% reduction) and BoundaryTPS retains 164 points (81/56/27; 79.5\% reduction). 

The clearest case where boundary emphasis helps is on the Circles datasets paired with linear SVM. Linear SVM is already poorly suited to concentric geometry (baseline G-Mean 0.484 on Circles Moderate Noise, 0.517 on Circles High Noise), and prototype selection magnifies the mismatch. Under TPS, linear SVM falls to 0.20 ($\Delta = -28.1$) on Circles Moderate Noise and 0.32 ($\Delta = -20.0$) on Circles High Noise. BoundaryTPS is less severe in both cases, with linear SVM at 0.35 ($\Delta = -13.6$) and 0.48 ($\Delta = -4.1$) respectively. On Two Moons (Moderate Noise), MLP classification under BoundaryTPS drops by $-23.0$ percentage points (TPS: $-12.0$) and 1-NN drops by $-11.5$ points (TPS: $-2.1$), despite the two methods retaining nearly identical fractions of the training data (15.0\% versus 15.1\%). Matched retention with diverging downstream performance points to BoundaryTPS over-resolving fine-scale boundary structure at the expense of interior support, and classifiers that depend on smooth interior decision surfaces (MLP in particular) are left without enough material to fit a clean boundary.

Taken together, the results show neither method dominating uniformly. TPS is more uniform across classifiers, as $\Delta$G-Mean is tightly clustered around zero on most datasets, and posts substantial gains on the imbalanced cases. BoundaryTPS trades some of that uniformity for a smaller computational footprint and tighter boundary preservation, which helps on some classifier–dataset pairings and hurts on others. Real-data evaluation in the next section is where the structural claims are tested.

\subsection{Performance on Real Data}
\label{sec:real_datasets}
 \begin{table}[tb]
\centering
\small
\begin{tabular}{lcccccccccc}
\hline\hline
\textbf{Dataset} & \textbf{n} & \textbf{d} & \textbf{Classes} & \textbf{IR} & \multicolumn{6}{c}{\textbf{Baseline G-Mean}} \\
\cline{6-11}
 & & & & & \textbf{1-NN} & \textbf{3-NN} & \textbf{5-NN} & \textbf{SVM} & \textbf{RF} & \textbf{MLP} \\
\hline
Iris              & 150  & 4  & 3  & 1.00  & 0.941 & 0.942 & 0.947 & 0.941 & 0.941 & 0.934 \\
Wine              & 178  & 13 & 3  & 1.48  & 0.957 & 0.957 & 0.974 & 0.961 & 0.990 & 0.978 \\
Sonar             & 208  & 60 & 2  & 1.14  & 0.868 & 0.830 & 0.801 & 0.742 & 0.813 & 0.877 \\
Ionosphere        & 351  & 34 & 2  & 1.79  & 0.803 & 0.767 & 0.755 & 0.855 & 0.919 & 0.888 \\
BreastCancer      & 569  & 30 & 2  & 1.68  & 0.943 & 0.957 & 0.961 & 0.971 & 0.951 & 0.969 \\
BloodTransfusion  & 748  & 4  & 2  & 3.20  & 0.507 & 0.544 & 0.560 & 0.023 & 0.506 & 0.562 \\
Diabetes          & 768  & 8  & 2  & 1.87  & 0.646 & 0.675 & 0.675 & 0.710 & 0.721 & 0.710 \\
MiceProtein       & 1080 & 77 & 8  & 1.43  & 0.998 & 0.993 & 0.984 & 0.995 & 0.999 & 0.998 \\
Digits            & 1797 & 64 & 10 & 1.05  & 0.973 & 0.975 & 0.979 & 0.980 & 0.978 & 0.979 \\
CardiotocographyA & 2126 & 35 & 10 & 10.93 & 0.997 & 0.997 & 0.997 & 1.000 & 1.000 & 0.999 \\
Ozone             & 2534 & 72 & 2  & 14.84 & 0.590 & 0.505 & 0.399 & 0.302 & 0.391 & 0.608 \\
KCvsKP            & 3196 & 73 & 2  & 1.09  & 0.922 & 0.952 & 0.944 & 0.969 & 0.992 & 0.994 \\
Spambase          & 4601 & 57 & 2  & 1.54  & 0.910 & 0.905 & 0.902 & 0.921 & 0.951 & 0.943 \\
Wilt              & 4839 & 5  & 2  & 17.54 & 0.757 & 0.716 & 0.681 & 0.641 & 0.868 & 0.936 \\
Satimage          & 6430 & 36 & 6  & 2.45  & 0.888 & 0.888 & 0.884 & 0.813 & 0.884 & 0.889 \\
\hline\hline
\\
\end{tabular}
\caption{Datasets used in the real Euclidean evaluation. Baseline G-Mean is the mean across 10 folds when training on the full dataset with no prototype selection.}
\label{tab:datasets}
\end{table}

We evaluate BoundaryTPS and TPS on fifteen real Euclidean datasets against CNN, CNN+ENN, DROP3, IB3, ICF, KMeans, and SPOTGreedy. Table~\ref{tab:datasets} lists each dataset with its size, dimensionality, class count, imbalance ratio, and the baseline G-Mean. We evaluate each method across topology preservation, selection stability, classification performance, class-imbalance preservation, and computational cost. 

\subsubsection{Topology Preservation}
\label{sec:topology_preservation}
We first examine whether each method's prototype set preserves the persistence diagram representation of the training set's topology. For each fold and each method, we compute the persistence diagram of the baseline training set $X_{\mathrm{train}}$ and of the corresponding prototype set $P$, and report the distance between the two diagrams. The Vietoris--Rips filtration is truncated at the minimum enclosing ball radius $r_{\mathrm{enc}}(X_{\mathrm{train}})$, which ensures finite distances and a scale comparable across datasets of differing diameter. 

Wasserstein-1 distance penalizes total off-diagonal displacement, while Wasserstein-2 distance squares those displacements and disproportionately punishes large birth--death shifts. Bottleneck distance reports only the worst single feature. The central metric used for comparison is the $H_1$ Wasserstein-1, on which loop structure dominates and the distinction between density-based and topology-aware selection is most informative. For completeness, we also evaluate the distance under the standard metrics of $H_0$ Wasserstein-1, $H_0$ and $H_1$ Wasserstein-2, and $H_1$ Bottleneck metrics, which allows the method's behavior to be characterized across the distinct quantities that the different distances penalize. 

BoundaryTPS achieves the lowest mean rank ($2.00$) across the $15$ datasets, with per-dataset wins on $6$ of the $15$ datasets (CardiotocographyA, Digits, Ionosphere, MiceProtein, Sonar, Wine). CNN is the next closest competitor at mean rank $3.20$, with $5$ per-dataset (BloodTransfusion, BreastCancer, Diabetes, Satimage, Spambase). TPS is third at $3.60$. The Friedman test rejects the null of equal ranks ($p = 6.82 \times 10^{-9}$), and the Nemenyi critical difference at $\alpha = 0.05$ is $\mathrm{CD} = 3.10$ implying BoundaryTPS is significantly more faithful in preserving the PD structure than DROP3 ($5.60$), CNN+ENN ($5.40$), ICF ($7.07$), SPOTGreedy ($7.00$), and KMeans ($6.87$). Per-dataset ranks supporting these aggregates are reported in the appendix (Table~\ref{tab:real_topology_per_dataset_rank}).

Table~\ref{tab:real_topology_combined} reports per-method mean Friedman ranks across all metrics and per-method magnitudes, normalized by the per-fold minimum enclosing ball radius to permit averaging across datasets of differing scale. The $H_1$ Wasserstein-1 result reproduces on $H_1$ Wasserstein-2, with BoundaryTPS at mean rank $2.87$ and Nemenyi-significant advantages over ICF, SPOTGreedy, and KMeans. The magnitude ordering on $H_1$ Wasserstein-1 places TPS ($7.215$), BoundaryTPS ($7.565$), and CNN ($7.813$) within $8\%$ of one another and ahead of the remaining methods. The slight reordering relative to the rank table (TPS first by magnitude, BoundaryTPS first by rank) reflects dataset heterogeneity in raw distance scale. BoundaryTPS achieves more first-place finishes, while TPS achieves smaller average distances. Both orderings support the same qualitative conclusion that the topology-aware methods cluster together at the top, separated by a visible gap from the lower-rank methods.

On the $H_1$ Bottleneck metric, the Friedman test fails to reject ($p = 0.22$), and we therefore do not draw rank-based conclusions from that column. On the two $H_0$ metrics, the ranking reorders substantially. CNN achieves mean ranks of $2.33$ and $2.27$ on $H_0$ Wasserstein-1 and Wasserstein-2 respectively, while BoundaryTPS sits at $5.00$ and $6.40$, with the $H_0$ Wasserstein-2 difference exceeding the Nemenyi critical difference against the column-best. 

The $H_0$ behavior does not reflect a failure to preserve connected-component structure, but rather reflects BoundaryTPS's filtration geometry. Because the vertex weights delay interior points, connected-component mergers occur at larger filtration values in the prototype diagram than in the baseline diagram. The components themselves are recovered as the prototype set is not topologically disconnected from the baseline, but their death times are displaced toward coarser scales. Wasserstein-1 sees the shifts linearly, predicting a milder middle-tier rank ($5.00$). Wasserstein-2 squares those displacements and is therefore the metric most punished by death-time shift, which predicts the bottom-tier rank ($6.40$).  

\begin{table}[tb]
\centering
\small
\begin{tabular}{lccccc}
\hline\hline
\textbf{Method} & \textbf{H$_1$ W$_1$} & \textbf{H$_1$ W$_2$} & \textbf{H$_0$ W$_1$} & \textbf{H$_0$ W$_2$} & \textbf{H$_1$ B} \\
\hline
BoundaryTPS & 2.00 & 2.87 & 5.00 & 6.40$^{\dagger}$ & 5.03 \\
 & \textit{(7.565)} & \textit{(0.251)} & \textit{(95.592)} & \textit{(2.872)} & \textit{(0.037)} \\
CNN & 3.20 & 2.93 & 2.33 & 2.27 & 3.70 \\
 & \textit{(7.813)} & \textit{(0.242)} & \textit{(85.793)} & \textit{(2.409)} & \textit{(0.036)} \\
TPS & 3.60 & 4.20 & 4.60 & 5.73$^{\dagger}$ & 5.27 \\
 & \textit{(7.215)} & \textit{(0.253)} & \textit{(96.269)} & \textit{(2.864)} & \textit{(0.037)} \\
IB3 & 4.27 & 4.07 & 3.60 & 3.60 & 3.80 \\
 & \textit{(7.867)} & \textit{(0.248)} & \textit{(87.974)} & \textit{(2.517)} & \textit{(0.037)} \\
CNN+ENN & 5.40$^{\dagger}$ & 5.47 & 6.00$^{\dagger}$ & 5.80$^{\dagger}$ & 5.07 \\
 & \textit{(8.553)} & \textit{(0.271)} & \textit{(101.655)} & \textit{(2.864)} & \textit{(0.038)} \\
DROP3 & 5.60$^{\dagger}$ & 5.33 & 5.67$^{\dagger}$ & 5.80$^{\dagger}$ & 4.93 \\
 & \textit{(8.505)} & \textit{(0.270)} & \textit{(100.638)} & \textit{(2.843)} & \textit{(0.038)} \\
KMeans & 6.87$^{\dagger}$ & 6.47$^{\dagger}$ & 6.87$^{\dagger}$ & 5.80$^{\dagger}$ & 5.60 \\
 & \textit{(8.789)} & \textit{(0.276)} & \textit{(106.169)} & \textit{(2.904)} & \textit{(0.037)} \\
SPOTGreedy & 7.00$^{\dagger}$ & 6.73$^{\dagger}$ & 5.93$^{\dagger}$ & 5.40$^{\dagger}$ & 6.03 \\
 & \textit{(8.802)} & \textit{(0.277)} & \textit{(105.066)} & \textit{(2.868)} & \textit{(0.038)} \\
ICF & 7.07$^{\dagger}$ & 6.93$^{\dagger}$ & 5.00 & 4.20 & 5.57 \\
 & \textit{(8.685)} & \textit{(0.276)} & \textit{(98.166)} & \textit{(2.696)} & \textit{(0.037)} \\
\hline
Friedman $p$ & 6.8e-09 & 4.7e-06 & 2.1e-04 & 2.8e-04 & 2.2e-01 \\
\hline\hline
\\
\end{tabular}
\caption{Per-method topology-preservation performance across all five metrics. The top number in each cell is the Friedman rank. Lower values are more faithful preservation of PD structure; a $\dagger$ marks methods whose rank exceeds the column-best by more than the Nemenyi critical difference at $\alpha = 0.05$. The italicized number in parentheses below is the magnitude of each per-fold distance divided by minimum enclosing ball radius $r_{\mathrm{enc}}(X_{\mathrm{train}})$. Rows sorted by H$_1$ W$_1$ rank. Friedman $p$ for each column is in the bottom row.}
\label{tab:real_topology_combined}
\end{table}

\begin{table}[tb]
\centering
\small
\begin{tabular}{lccc}
\hline\hline
\textbf{Method} & \textbf{Betti H$_0$} & \textbf{Betti H$_1$} & \textbf{Euler} \\
\hline
BoundaryTPS & 3.93 & 2.27 & 3.80 \\
CNN & 2.53 & 2.80 & 2.67 \\
TPS & 4.27 & 3.80 & 4.20 \\
IB3 & 4.00 & 3.67 & 4.00 \\
CNN+ENN & 6.27$^{\dagger}$ & 6.40$^{\dagger}$ & 6.40$^{\dagger}$ \\
DROP3 & 5.73$^{\dagger}$ & 5.73$^{\dagger}$ & 5.73 \\
KMeans & 7.07$^{\dagger}$ & 6.53$^{\dagger}$ & 7.13$^{\dagger}$ \\
SPOTGreedy & 6.20$^{\dagger}$ & 6.87$^{\dagger}$ & 6.13$^{\dagger}$ \\
ICF & 5.00 & 6.93$^{\dagger}$ & 4.93 \\
\hline
Friedman $p$ & 5.6e-05 & 4.0e-09 & 4.6e-05 \\
\hline\hline
\\
\end{tabular}
\caption{Per-method mean Friedman rank across Betti number curves $\beta_0(r)$ and $\beta_1(r)$, and the Euler characteristic curve $\chi(r) = \beta_0(r) - \beta_1(r)$. Curves are computed on a uniform grid over $[0, r_{\mathrm{enc}}(X_{\mathrm{train}})]$ with the minimum enclosing radius as the truncation, and distances between baseline and prototype curves are reported in the $L_2$ function-space norm. For each column, $\dagger$ marks methods whose mean rank exceeds the column-best by more than the Nemenyi critical difference at $\alpha = 0.05$. Friedman $p$ for each column is in the bottom row.}
\label{tab:real_topology_betti_euler_summary}
\end{table}

To verify this, we computed $L_2$ distances between baseline and prototype Betti-number curves $\beta_0(r)$, $\beta_1(r)$, and the Euler-characteristic curve $\chi(r) = \beta_0(r) - \beta_1(r)$ on a uniform grid over $[0, r_{\mathrm{enc}}]$ for both $H_0$ and $H_1$ homology groups. Table~\ref{tab:real_topology_betti_euler_summary} provides the Friedman ranks. The three metrics form a coherent ordering under the death-time-shift (for $H_0$, $W_2 > W_1 > L_2$ in rank), supporting the interpretation that BoundaryTPS loses death-time \textit{coordinate} fidelity on $H_0$ Wasserstein distances, not the components themselves. 

The same argument does not apply to $H_1$ as loops are not displaced by a death-time shift. They are either preserved or destroyed, and a density-based selection that fails to sample around a loop's circumference removes the feature outright. On Betti $H_1$, BoundaryTPS achieves the lowest mean rank ($2.27$) with Nemenyi-significant advantages over $5$ of the competing methods (Friedman $p = 4.0 \times 10^{-9}$). 

\begin{figure}
    \centering
    \includegraphics[width=\linewidth]{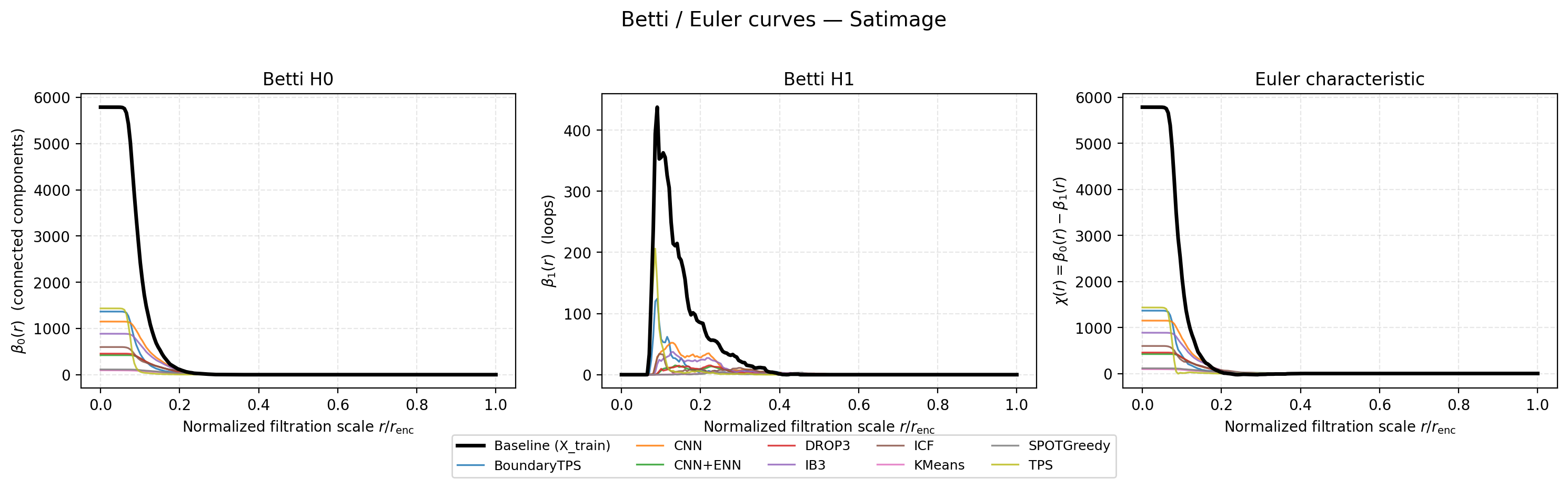}
    \caption{Betti and Euler-characteristic curves on Satimage. Panels show $\beta_0(r)$, the number of connected components (left); $\beta_1(r)$, the number of loops (middle); and the Euler characteristic $\chi(r) = \beta_0(r) - \beta_1(r)$ (right). All three are evaluated on a uniform grid over the normalized filtration scale $r$/$r_{enc}$, where $r_{enc}$ is the minimum enclosing ball radius of the training set. The baseline curves (black) are computed on $X_{train}$ with the colored curves for each method's prototype set. BoundaryTPS (blue) and TPS (yellow) are the only methods that recover a discernible $\beta_1$ peak in roughly the same scale window as the baseline; the remaining methods flatten the H1 curve nearly to zero.}
    \label{fig:satimage_betti_euler}
\end{figure}

Figure~\ref{fig:satimage_betti_euler} displays an example of the underlying baseline and prototype curves for Satimage and visualizes the recovery directly. We provide the remaining dataset curves in the Appendix. Changing the truncation to $2 \times r_{\mathrm{enc}}$ for sensitivity produces identical orderings and is reported in the supplementary materials.

\subsubsection{Selection Stability}
\label{sec:selection_stability}

Next we look at the reproducibility of the selected index set across re-samplings of the training data. We measure stability as the mean pairwise Jaccard similarity of prototype index sets across the $10$ stratified outer folds, averaged across the $15$ real datasets. Table~\ref{tab:real_jaccard_stability} reports the result. Note, KMeans is excluded because its synthetic centroids have no index space against which Jaccard is defined.

TPS achieves the highest mean Jaccard similarity ($0.225$), followed by BoundaryTPS ($0.202$). The closest classical competitor is CNN at $0.170$ with ICF ($0.157$) and IB3 ($0.134$) following. DROP3 ($0.080$), SPOTGreedy ($0.076$), and CNN+ENN ($0.057$) sit substantially lower, with the latter at roughly one-quarter of TPS's stability. Selection criteria depending on chained local decisions produce a compounded instability under fold perturbation, while criteria depending on a single decision rule (CNN's misclassified-instance retention) or on global structural features (the persistence-based selection of TPS and BoundaryTPS) are more reproducible. 

CNN+ENN, DROP3, and ICF all combine an edition step with a second selection rule, and each link in the chain is a potential source of instability that the table reflects. TPS defines filtrations globally over a class, and substituting a small fraction of training points typically perturbs only the boundaries of the persistence diagram, not which features end up selected.

\begin{table}[tb]
\centering
\small
\begin{tabular}{lcc}
\hline\hline
\textbf{Method} & \textbf{Mean Jaccard} (across datasets) & \textbf{Std} \\
\hline
TPS & 0.225 & 0.115 \\
BoundaryTPS & 0.202 & 0.114 \\
CNN & 0.170 & 0.106 \\
ICF & 0.157 & 0.108 \\
IB3 & 0.134 & 0.115 \\
DROP3 & 0.080 & 0.042 \\
SPOTGreedy & 0.076 & 0.072 \\
CNN+ENN & 0.057 & 0.025 \\
\hline\hline
\\
\end{tabular}
\caption{Average pairwise Jaccard similarity of the selected index sets across the 10 stratified CV folds, then averaged across real-world datasets. Higher values mean more reproducible.}
\label{tab:real_jaccard_stability}
\end{table}

\subsubsection{Classification Performance}
\label{sec:classification_performance}
\begin{table}[tb]
\centering
\small
\begin{tabular}{lccccccc}
\hline\hline
\textbf{Method} & \textbf{1-NN} & \textbf{3-NN} & \textbf{5-NN} & \textbf{Lin.\ SVM} & \textbf{MLP} & \textbf{RF} & \textbf{Red.\ (\%)} \\
\hline
SPOTGreedy & 0.844 & 0.820 & 0.803 & 0.852 & 0.877 & 0.858 & 91.8 \\
 & $(-0.3)$ & $(-2.0)$ & $(-2.6)$ & $(+6.3)$ & $(-0.8)$ & $(-0.2)$ & \\[3pt]
KMeans & 0.857 & 0.809 & 0.782 & 0.863 & 0.885 & 0.843 & 92.0 \\
 & $(+1.1)$ & $(-3.1)$ & $(-4.7)$ & $(+7.4)$ & $(+0.0)$ & $(-1.7)$ & \\[3pt]
DROP3 & 0.820 & 0.814 & 0.809 & 0.838 & 0.850 & 0.814 & 89.5 \\
 & $(-2.7)$ & $(-2.6)$ & $(-2.0)$ & $(+4.9)$ & $(-3.4)$ & $(-4.6)$ & \\[3pt]
CNN & 0.836 & 0.787 & 0.710 & 0.810 & 0.869 & 0.837 & 77.4 \\
 & $(-1.1)$ & $(-5.3)$ & $(-11.9)$ & $(+2.2)$ & $(-1.5)$ & $(-2.3)$ & \\[3pt]
TPS & 0.789 & 0.788 & 0.762 & 0.813 & 0.822 & 0.778 & 77.4 \\
 & $(-5.8)$ & $(-5.3)$ & $(-6.8)$ & $(+2.4)$ & $(-6.2)$ & $(-8.2)$ & \\[3pt]
BoundaryTPS & 0.757 & 0.752 & 0.739 & 0.762 & 0.804 & 0.759 & 81.7 \\
 & $(-8.9)$ & $(-8.8)$ & $(-9.1)$ & $(-2.6)$ & $(-8.0)$ & $(-10.1)$ & \\[3pt]
IB3 & 0.768 & 0.640 & 0.571 & 0.725 & 0.798 & 0.712 & 79.9 \\
 & $(-7.9)$ & $(-20.1)$ & $(-25.9)$ & $(-6.3)$ & $(-8.6)$ & $(-14.8)$ & \\[3pt]
ICF & 0.639 & 0.575 & 0.548 & 0.752 & 0.824 & 0.790 & 86.9 \\
 & $(-20.8)$ & $(-26.5)$ & $(-28.1)$ & $(-3.7)$ & $(-6.1)$ & $(-7.0)$ & \\[3pt]
CNN+ENN & 0.575 & 0.467 & 0.398 & 0.542 & 0.635 & 0.537 & 91.7 \\
 & $(-27.1)$ & $(-37.3)$ & $(-43.1)$ & $(-24.7)$ & $(-24.9)$ & $(-32.3)$ & \\
\hline\hline
\\
\end{tabular}
\caption{Per-method mean G-Mean across all real-world datasets, broken out by classifier, with mean $\Delta$G-Mean (percentage points relative to the baseline) shown beneath each value. \textbf{Red.\ (\%)} reports the mean reduction percentage across datasets and folds.}
\label{tab:real_per_method_gmean}
\end{table}

Table~\ref{tab:real_per_method_gmean} reports mean G-Mean across the fifteen real-world datasets, broken out by classifier. The leading methods on aggregate G-Mean are SPOTGreedy ($0.84$) and KMeans ($0.84$), followed by DROP3 ($0.82$) and CNN ($0.81$). TPS ($0.79$) and BoundaryTPS ($0.76$) sit in the middle of the field, ahead of IB3 ($0.70$), ICF ($0.69$), and CNN+ENN ($0.53$). IB3 collapses on the $k$-NN classifiers in particular, with $\Delta$G-Mean of $-20.1$ and $-25.9$ percentage points on 3-NN and 5-NN, while CNN+ENN degrades on every classifier.

Under classification performance, TPS and BoundaryTPS are seemingly indistinguishable. Friedman ranks differ by at most $1.20$ between TPS and BoundaryTPS across the six classifiers, and neither method is flagged as significantly different from the other in the Bonferroni-Dunn tables. In G-Mean terms, choosing between the methods does not change downstream outcomes by a margin the present data can resolve, and the practical case for choosing one over the other rests on the other properties documented. 

Bonferroni-Dunn pairwise tests with TPS / BoundaryTPS as the relevant control are presented as separate tables in the appendix. TPS is significantly worse than KMeans on 1-NN and MLP and than CNN on MLP, but significantly better than ICF on 3-NN and 5-NN and than CNN+ENN on the same two classifiers. BoundaryTPS shows a wider gap to the top-tier methods (significantly worse than KMeans, CNN, and SPOTGreedy across multiple classifiers) but remains significantly better than CNN+ENN on 5-NN.

\begin{table}[tb]
\centering
\small
\begin{tabular}{lcccccc}
\hline\hline
\textbf{Method} & \textbf{1-NN} & \textbf{3-NN} & \textbf{5-NN} & \textbf{Lin.\ SVM} & \textbf{MLP} & \textbf{RF} \\
\hline
CNN & 21.3 & 15.3 & 13.3 & 28.7 & 19.3 & 16.7  \\
 & $(55.3)$ & $(50.7)$ & $(50.0)$ & $(48.0)$ & $(55.3)$ & $(53.3)$  \\[3pt]
DROP3 & 10.7 & 11.3 & 14.7 & 20.0 & 14.0 & 8.7  \\
 & $(30.0)$ & $(40.0)$ & $(56.7)$ & $(42.0)$ & $(29.3)$ & $(32.0)$ \\[3pt]
KMeans & 26.7 & 19.3 & 16.0 & 34.0 & 30.0 & 14.7  \\
 & $(64.7)$ & $(40.0)$ & $(40.0)$ & $(58.0)$ & $(54.7)$ & $(40.0)$ \\[3pt]
SPOTGreedy & 22.7 & 19.3 & 18.0 & 24.7 & 25.3 & 26.7 \\
 & $(40.7)$ & $(41.3)$ & $(38.0)$ & $(48.0)$ & $(50.7)$ & $(40.0)$  \\[3pt]
IB3 & 4.7 & 4.0 & 4.0 & 17.3 & 18.7 & 9.3  \\
 & $(24.0)$ & $(18.7)$ & $(17.3)$ & $(29.3)$ & $(42.7)$ & $(32.7)$ \\[3pt]
TPS & 7.3 & 11.3 & 8.7 & 18.7 & 9.3 & 12.7  \\
 & $(24.7)$ & $(30.0)$ & $(28.7)$ & $(34.0)$ & $(16.0)$ & $(20.7)$ \\[3pt]
BoundaryTPS & 6.7 & 8.7 & 7.3 & 16.7 & 8.7 & 11.3  \\
 & $(14.7)$ & $(20.7)$ & $(18.0)$ & $(22.7)$ & $(16.0)$ & $(18.0)$ \\[3pt]
ICF & 6.7 & 6.0 & 4.7 & 11.3 & 12.0 & 11.3  \\
 & $(11.3)$ & $(9.3)$ & $(6.7)$ & $(20.0)$ & $(22.0)$ & $(28.7)$ \\[3pt]
CNN+ENN & 2.7 & 3.3 & 0.7 & 4.0 & 6.7 & 1.3  \\
 & $(6.7)$ & $(7.3)$ & $(3.3)$ & $(10.7)$ & $(10.0)$ & $(4.7)$  \\
\hline\hline
\\
\end{tabular}
\caption{Per-fold rank distribution across all (dataset, fold) instances. The upper row per method reports the percent of combinations of (datasets, fold, classifier) in which the method was rank 1 (\textbf{top-1\%}), and the lower row in parentheses reports the percent in which it was rank 3 or better (\textbf{top-3\%}).}
\label{tab:real_perfold_rank_distribution}
\end{table}

Aggregate $\Delta$G-Mean rankings mask fold-level variability. Table~\ref{tab:real_perfold_rank_distribution} reports the per-fold rank distribution per method. TPS reaches rank-$1$ G-Mean in $11.3\%$ of (dataset, fold, classifier) combinations and rank-$3$-or-better in $25.7\%$ on average. BoundaryTPS reaches rank-$1$ in $9.9\%$ on average and rank-$3$-or-better in $18.4\%$ on average. These rank-$1$ frequencies show that the methods are not uniformly dominated by classical methods, and are capable of producing favorable prototype sets on a subset of (dataset, method, fold) combinations. Building variants that more uniformly improve discriminative classification without sacrificing the topology-preservation property is an open direction.

% ============================================================================

\subsubsection{Class-Imbalance Preservation}
\label{sec:class_imbalance_behavior}

BoundaryTPS ($1.30$) and TPS ($1.59$) achieve the smallest mean deviations because neither selection rule conditions on class labels. The persistence-feature-based selection retains points in proportion to their topological participation, which inherits the source set's class proportions as an artifact. KMeans and SPOTGreedy reach identical mean deviations of $3.20$ (with identical standard deviations of $5.305$), because both produce perfectly balanced prototype sets in every fold via per-class budgeting. IB3 is a separate case as its mean deviation of $16.04$ and standard deviation of $52.25$ against a median of $0.94$ indicate that IB3 typically produces a prototype set close in IR to the source, but occasionally collapses one class entirely on adversarial folds, producing the long-tailed distribution that drives the mean. The remaining methods sit at intermediate mean deviations between $2.87$ (DROP3) and $3.61$ (ICF), with label-aware selection rules that neither enforce $\mathrm{IR} = 1$ nor preserve the source IR. Full per-method statistics are in the appendix (Table~\ref{tab:real_ir_deviation}).

Whether IR preservation or IR enforcement is preferable depends on downstream goals, such as classification or dataset compression that does not disrupt underlying characteristics for storage. As such, we offer this analysis as a potential point of interest, but make no comparative claim.

\subsubsection{Computational Cost}
\label{sec:computational_cost}

\begin{figure}
    \centering
    \includegraphics[width=0.75\linewidth]{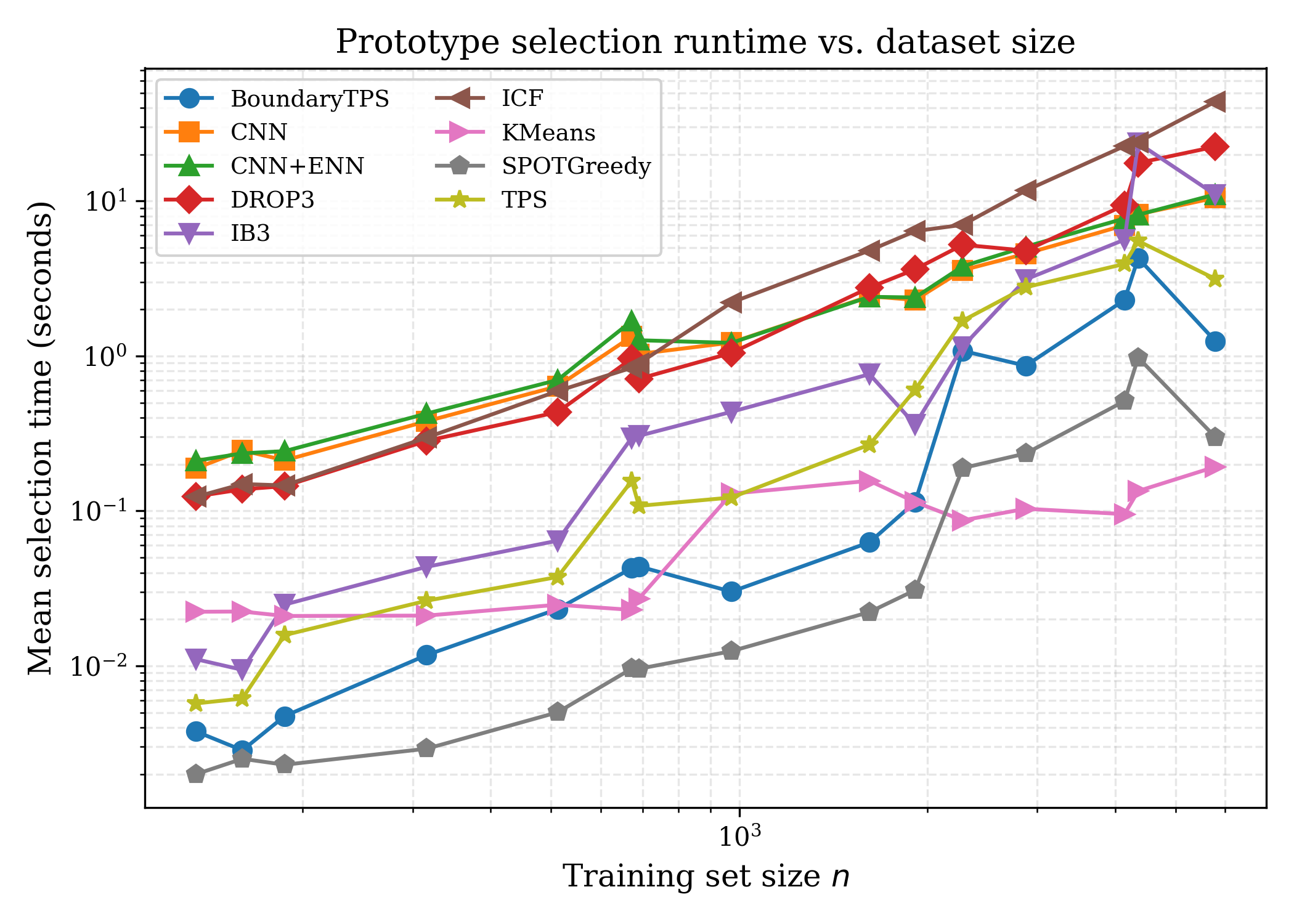}
    \caption{Mean prototype-selection runtime (seconds) as a function of training-set size $n$, plotted on log–log axes. Each point is the average wall-clock selection time across the real datasets; all methods were run on identical hardware with a single thread.}
    \label{fig:selection_time}
\end{figure}

Figure~\ref{fig:selection_time} reports mean prototype selection time as a function of training set size on log-log axes for the evaluated methods. Selection time scales approximately linearly on the log-log axes for every method, consistent with polynomial scaling in $n$ over the range of dataset sizes tested.

At the largest dataset Satimage, the methods separate cleanly. KMeans ($0.20$s) and SPOTGreedy ($0.29$s) are the fastest, followed by BoundaryTPS ($1.25$s) and TPS ($2.59$s). The chained-rule methods CNN ($10.15$s), CNN+ENN ($10.89$s), and IB3 ($11.08$s) sit roughly an order of magnitude slower, with DROP3 ($22.15$s) and ICF ($44.66$s) the slowest. The cost advantage of TPS and BoundaryTPS over the chained-rule methods is consistent across the dataset-size range and widens as $n$ grows. At $n \geq 4{,}000$, BoundaryTPS completes in the $1$--$5$ second range and TPS in the $2$--$6$ second range, against tens of seconds for the slowest classical methods. This margin supports deployment of both topological methods at training set scales where the chained-rule methods become impractical, despite the combinatorial nature of simplicial complex filtrations.

Log-log regression of selection time on $n_{\mathrm{train}}$ yields empirical scaling exponents of $1.84$ for TPS ($R^2 = 0.94$) and $1.87$ for BoundaryTPS ($R^2 = 0.91$), consistent with the $\mathcal{O}(n^2 d)$ conjectured bound and reflecting GUDHI's pruning of Rips edges above the filtration threshold. 

\subsection{Ablation Studies} \label{section:ablation}
We conduct three ablation studies to characterize the behavior of TPS and BoundaryTPS along axes the main evaluation does not isolate with their (i) sensitivity to method hyperparameters, (ii) the effect of homology dimension on the selected prototype set, and (iii) robustness to label noise. The third study includes all competing methods, while the first two are specific to the topological methods. We use 1-NN to focus on geometric properties of the prototype set rather than classifier-specific behavior. As such, we do not claim ablation conclusions transfer uniformly to other downstream classifiers.

\subsubsection{Hyperparameter Sensitivity}

We sweep each hyperparameter for TPS and BoundaryTPS individually while holding the others at default values, isolating the marginal effect on G-Mean and reduction percentage. Figures~\ref{fig:hp_tps} and~\ref{fig:hp_btps} in the appendix show the results, averaged across the nine simulated datasets.

For TPS, the quantile parameter $q$ is the dominant lever. At $q = 0.01$, TPS retains only the shortest-lived persistence features, yielding $97.6\%$ reduction at the cost of mean G-Mean dropping to $0.682$. As $q$ increases, reduction declines monotonically from $90.1\%$ at $q = 0.05$ to $34.8\%$ at $q = 0.50$, while G-Mean rises sharply to a plateau near $0.79$ for $q \geq 0.10$ and remains stable through $q = 0.50$. The plateau is consistent with classification-relevant topological structure being concentrated in moderate-lifetime features. Longer-lived features add prototypes without changing downstream classification appreciably. The neighborhood parameter $k$ has comparatively little effect as mean G-Mean varies by only $0.003$ across the full range ($k = 1$ to $k = 20$), and reduction by less than one percentage point. TPS therefore behaves as a single-hyperparameter method in practice.

For BoundaryTPS, the picture is reversed. Hyperparameter $k$ is the dominant parameter and the most fragile. At $k = 1$, BoundaryTPS reaches mean G-Mean $0.759$. Increasing to $k = 3$ drops G-Mean to $0.523$, and $k = 20$ leaves it at $0.431$. Because $k$ enters the vertex-weight definition directly, larger $k$ averages each weight over more distant non-target neighbors, which is consistent with a diluted boundary signal and weights that fail to separate boundary from interior points. The degradation concentrates on datasets with complex boundary geometry.

The quantile parameter $q$ in BoundaryTPS behaves differently than in TPS. G-Mean rises monotonically from $0.371$ at $q = 0.01$ to $0.684$ at $q = 0.50$, a pattern consistent with the weighted filtration already concentrating topological information at boundary-proximate vertices. Longer-lived features capture broader boundary structure rather than introducing interior redundancy. The tolerance parameter $\tau$ has minimal effect on G-Mean (range $0.513$--$0.546$ across $\tau \in [0, 0.3]$) but reduces selection aggressiveness from $83.2\%$ at $\tau = 0$ to $59.9\%$ at $\tau = 0.3$, since a wider tolerance window admits more features into the prototype set.

\begin{figure}[tb]
    \centering
    \includegraphics[width=.75\linewidth]{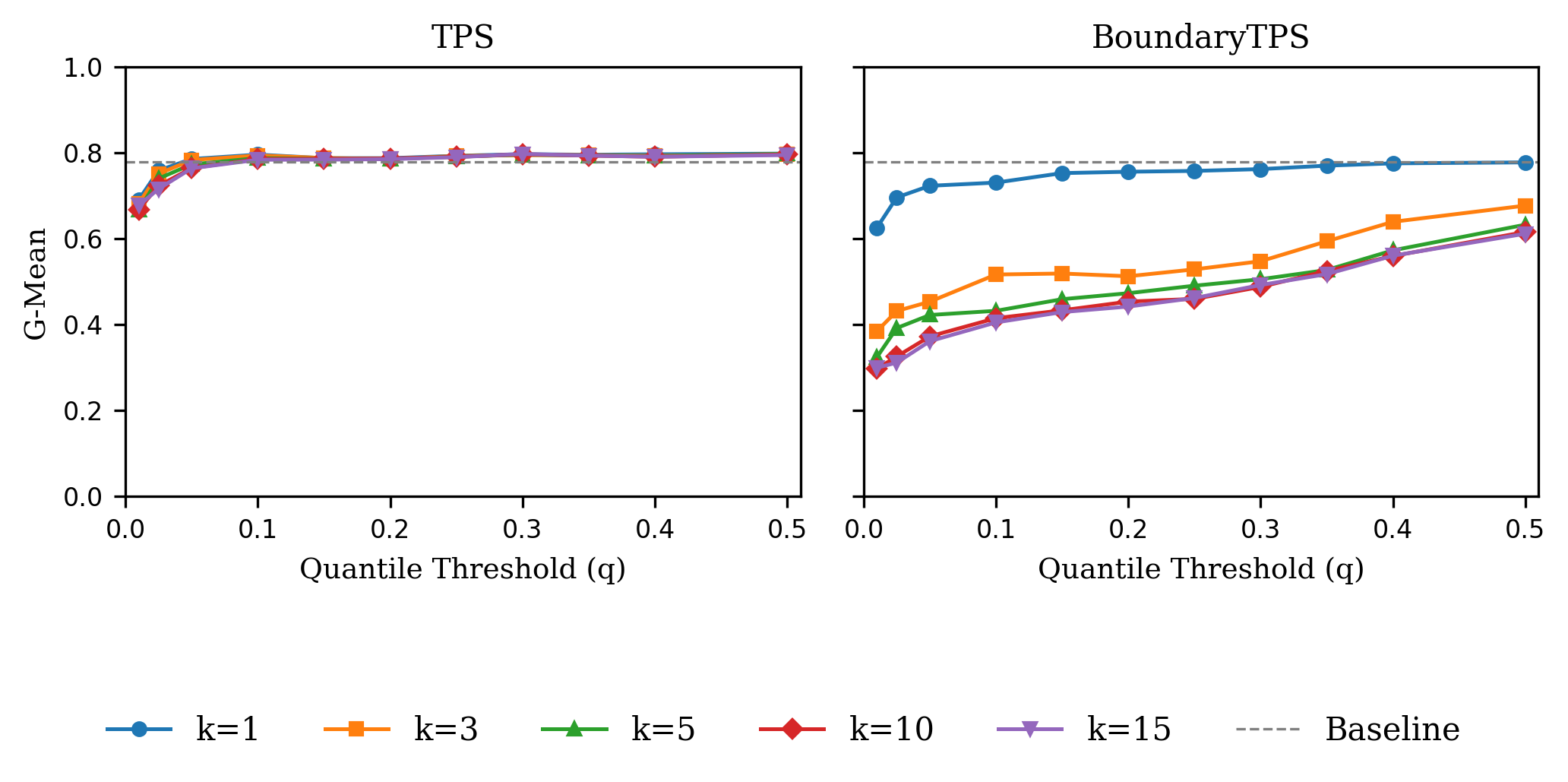}
    \caption{Sensitivity of TPS (left) and BoundaryTPS (right) to the quantile parameter $q$ across neighborhood sizes $k$, averaged across the nine simulated datasets under 1-NN. The dashed line indicates the baseline. TPS shows minimal sensitivity to $q$ for $q \geq 0.025$; BoundaryTPS degrades sharply at low quantiles for $k > 1$.}
    \label{fig:q_sensitivity}
\end{figure}

To examine the joint effect of $q$ and $k$, we sweep $q \in \{0.01, 0.025, 0.05, 0.10, \ldots, 0.50\}$ at $k \in \{1, 3, 5, 10, 15\}$ and report mean G-Mean and reduction averaged across the nine simulated datasets under 1-NN. Figure~\ref{fig:q_sensitivity} shows the resulting tradeoff curves.

For TPS, the curves are nearly identical across all values of $k$, confirming the marginal insensitivity observed in the univariate sweep. BoundaryTPS shows a fundamentally different pattern. At $k = 1$, BoundaryTPS reaches its best G-Mean of $0.697$ at $q = 0.05$ with $91.4\%$ reduction. At $k \geq 3$ the best achievable G-Mean collapses to $0.539$ at $k = 3$, $0.441$ at $k = 5$, and below $0.430$ for $k \geq 10$, with the optimum shifting to $q = 0.50$ (the least aggressive selection). BoundaryTPS's vertex weights delay interior points in the filtration, so at low quantiles the prototype set concentrates in a narrow boundary shell, and as $k$ grows that shell is computed against more distant non-target neighbors and can exclude minority instances entirely. The interaction coupled with the ablation on $\tau$ also confirms BoundaryTPS as a one-hyperparameter method in practice. $k$ must be held at $1$ and $\tau = 0$ to remain competitive, with $q$ providing the remaining degrees of freedom in tuning.

\subsubsection{Effect of Homology Dimension on Prototype Composition}

Including higher-dimensional topological features should plausibly enlarge the prototype set, since closing a $1$-cycle requires participation from more vertices than identifying a connected component. We compute prototype sets under $H_0$, $H_1$, and the union $H_0 \cup H_1$ for TPS and BoundaryTPS on all nine simulated datasets. Table~\ref{tab:homology_ablation} reports prototype counts and reduction percentages for each configuration, reporting set composition only.

For TPS, $H_0$ retains substantially fewer prototypes than $H_1$ across every dataset, with mean reduction of $60.4\%$ under $H_0$ versus $10.1\%$ under $H_1$. The union $H_0 \cup H_1$ adds zero vertices beyond $H_1$ on 8 of 9 datasets with an average of $0.3$ vertices on Mixed Multiclass. The near-complete containment follows from the simplex-enumeration proxy remark in Section 3.3. The $(p+1)$-simplices that enter an unweighted Rips filtration up to the $H_1$ truncation include every edge that enters up to the $H_0$ truncation, so the $H_1$ vertex set returned by the proxy is essentially a superset of the $H_0$ vertex set. This is a property of the recovery mechanism, not an algebraic identity between the homology groups themselves. As consequence, $H_1$ acts closer to a noise edition method instead of an aggressive prototype selection method. 

For BoundaryTPS, the relationship is more complex. $H_0$ retains $68.0\%$ mean reduction with a tight range across datasets ($45.6$--$72.2\%$) while the union $H_0 \cup H_1$ introduces extra vertices on 6 of 9 datasets, averaging $12$ extra vertices per fold. The extras concentrate on datasets with complex boundary geometry such as Mixed Multiclass, Concentric Circles (High), and Two Moons (High). The non-containment indicates that BoundaryTPS's weighted filtration produces an $H_1$ landscape distinct from TPS's unweighted Rips filtration. Under boundary-weighted distances, the vertices participating in long-lived $1$-cycles are not always a superset of those participating in long-lived connected-component features, and on geometrically complex datasets the two sets identify partially disjoint regions.

The non-containment pattern is concentrated on the same datasets that drove BoundaryTPS's $k$-sensitivity. The convergence of the two is suggestive rather than diagnostic. Both effects involve the weighted filtration and complex boundary geometry, but we do not isolate a shared mechanism here. This motivates a unified examination of weighted-filtration stability on complex geometries as a direction for future work.

\begin{table}[tb]
\centering
\caption{Homology ablation: prototype counts by dimension and extra vertices contributed by $H_1$ in the union $H_0 \cup H_1$ (mean across 10 folds, 1-NN downstream). Extra refers to additional vertices that appear in the union $H_0 \cup H_1$ that are not in $\max{|H_0|, |H_1|}$. } 
\label{tab:homology_ablation}
\begin{tabular}{c l c c c c c}
\hline\hline
\textbf{Method} & \textbf{Dataset} & $n_{\mathrm{train}}$ & $|H_0|$ & $|H_1|$ & $|H_0 \cup H_1|$ & \textbf{Extra} \\
\hline
TPS & Well-Sep.\ Blobs     & 540 & 226 & 476 & 476 & 0 \\
TPS & Overlap.\ Blobs      & 720 & 255 & 619 & 619 & 0 \\
TPS & Overlap.\ Clusters   & 540 & 187 & 480 & 480 & 0 \\
TPS & Two Moons (Mod.)     & 450 & 208 & 419 & 419 & 0 \\
TPS & Two Moons (High)     & 450 & 166 & 402 & 402 & 0 \\
TPS & Circles (Mod.)       & 450 & 216 & 428 & 428 & 0 \\
TPS & Circles (High)       & 450 & 157 & 409 & 409 & 0 \\
TPS & Imbalanced (80/20)   & 450 & 180 & 401 & 401 & 0 \\
TPS & Mixed Multiclass     & 720 & 279 & 633 & 634 & 1 \\
\hline
BoundaryTPS & Well-Sep.\ Blobs    & 540 & 294 & 540 & 540 & 0 \\
BoundaryTPS & Overlap.\ Blobs     & 720 & 208 & 145 & 236 & 17 \\
BoundaryTPS & Overlap.\ Clusters  & 540 & 160 & 200 & 246 & 18 \\
BoundaryTPS & Two Moons (Mod.)    & 450 & 146 & 450 & 450 & 0 \\
BoundaryTPS & Two Moons (High)    & 450 & 125 & 305 & 336 & 21 \\
BoundaryTPS & Circles (Mod.)      & 450 & 136 & 51  & 136 & 0 \\
BoundaryTPS & Circles (High)      & 450 & 127 & 120 & 150 & 22 \\
BoundaryTPS & Imbalanced (80/20)  & 450 & 128 & 442 & 444 & 2 \\
BoundaryTPS & Mixed Multiclass    & 720 & 203 & 553 & 582 & 30 \\
\hline\hline
\end{tabular}
\end{table}

\subsubsection{Label Noise Robustness}
We inject symmetric label noise at rates $\eta \in \{0\%, 10\%, 15\%, 20\%\}$ into three simulated datasets spanning a range of class overlap (Well-Separated Blobs, Overlapping Blobs, Overlapping Clusters), reassigning a fraction $\eta$ of training labels to a uniformly random alternative class, and measure the resulting G-Mean under 1-NN. Table~\ref{tab:noise_robustness} reports mean G-Mean across the three datasets at each noise level.

TPS absolute G-Mean drops by $0.090$ from $\eta = 0\%$ ($0.714$) to $\eta = 20\%$ ($0.624$), but TPS outperforms the baseline at every noise level, with the gap widening with $\eta$. The results suggest TPS's two-stage construction produces noise-filtering behavior, despite not being designed explicitly for that purpose. A plausible mechanism is that flipped labels produce within-class topological irregularities that the second-stage within-class filtration prunes disproportionately, so the prototype set sees a lower effective noise rate than the input.

BoundaryTPS is the least noise-robust method in the comparison, degrading by $0.213$ ($0.568 \to 0.355$) and falling well below the baseline at every $\eta > 0\%$. A class point whose label has flipped into a nontarget-region neighborhood receives a low weight (because its $k$ nearest neighbors are now ``non-target'' under its corrupted label), and the filtration treats it as a genuine boundary point. We note that the failure under label noise is shared with the generation methods, which select centroids or representative points without any consistency check. 

Localizing topological information at boundary regions is protective when the boundary is correctly identified, but label noise corrupts the very signal BoundaryTPS uses to identify the boundary, so the method's design amplifies rather than damps the perturbation. We treat this as a scope limitation of BoundaryTPS rather than of topological prototype selection generally as TPS's two-stage construction, which retains an interior-typical sample independent of boundary weights, provides a working pathway to noise robustness within the same framework.

\begin{table}[tb]
\centering
\caption{Mean G-Mean ($\pm$ std) under label noise, averaged across three simulated datasets (Well-Separated Blobs, Overlapping Blobs, Overlapping Clusters) with 1-NN classifier.}
\label{tab:noise_robustness}
\begin{tabular}{lcccc}
\hline\hline
\textbf{Method} & $\eta=0\%$ & $\eta=10\%$ & $\eta=15\%$ & $\eta=20\%$ \\
\hline 
Baseline & $0.703 \pm 0.227$ & $0.638 \pm 0.197$ & $0.605 \pm 0.189$ & $0.582 \pm 0.175$ \\
\hline
DROP3        & $0.717 \pm 0.229$ & $0.716 \pm 0.227$ & $0.713 \pm 0.223$ & $0.711 \pm 0.227$ \\
CNN+ENN      & $0.722 \pm 0.232$ & $0.711 \pm 0.229$ & $0.695 \pm 0.228$ & $0.673 \pm 0.238$ \\
TPS          & $0.714 \pm 0.225$ & $0.672 \pm 0.194$ & $0.648 \pm 0.186$ & $0.624 \pm 0.176$ \\
CNN          & $0.690 \pm 0.236$ & $0.608 \pm 0.189$ & $0.562 \pm 0.178$ & $0.554 \pm 0.161$ \\
SPOTGreedy   & $0.691 \pm 0.241$ & $0.592 \pm 0.184$ & $0.546 \pm 0.156$ & $0.520 \pm 0.163$ \\
KMeans       & $0.688 \pm 0.234$ & $0.572 \pm 0.158$ & $0.534 \pm 0.143$ & $0.507 \pm 0.137$ \\
IB3          & $0.669 \pm 0.249$ & $0.574 \pm 0.177$ & $0.522 \pm 0.144$ & $0.509 \pm 0.147$ \\
ICF          & $0.636 \pm 0.269$ & $0.587 \pm 0.248$ & $0.563 \pm 0.204$ & $0.505 \pm 0.171$ \\
BoundaryTPS  & $0.568 \pm 0.315$ & $0.413 \pm 0.191$ & $0.390 \pm 0.155$ & $0.355 \pm 0.152$ \\
\hline\hline
\end{tabular}
\end{table}

\section{Discussions and Future Work}
\label{section:conclusion}
The empirical evaluation identifies several structural properties of topological prototype selection. We summarize and characterize the mechanisms and limitations of TPS and BoundaryTPS and close with future research directions.

\subsection{Discussion}
The headline result is the topology-preservation property of the proposed methods. BoundaryTPS achieves the lowest mean Friedman rank ($2.00$) on $H_1$ Wasserstein-1 distance across the fifteen real datasets, winning $6$ of the $15$ datasets outright, with TPS third at $3.60$. The Friedman test rejects equality at $p = 6.82 \times 10^{-9}$, and the Nemenyi critical difference at $\alpha = 0.05$ ($\mathrm{CD} = 3.10$) places BoundaryTPS significantly ahead of DROP3, CNN+ENN, ICF, SPOTGreedy, and KMeans for $H_1$ persistence-diagram preservation. The $H_0$ ranking is different, and we read that difference as a death-time-shift effect rather than a failure to preserve connected components. BoundaryTPS's vertex weights delay interior points, meaning  the components survive, but merges happen at larger filtration values in the prototype diagram than in the baseline diagram. The magnitude of the rank penalty under Wasserstein-2 (6.40), Wasserstein-1 (5.00), and the Betti-$H_0$ curve metric (3.93) give the empirical evidence.

The stability and class-imbalance behaviors of the proposed methods follow from the same global-versus-local distinction. TPS achieves the highest mean Jaccard ($0.225$) and BoundaryTPS the second-highest ($0.202$). By contrast, selection rules depending on chained local decisions (DROP3 at $0.080$, SPOTGreedy at $0.076$, CNN+ENN at $0.057$) exhibit compound instability under fold perturbation that the topology-based methods do not. The class-imbalance behavior is a similar consequence of construction. Because neither TPS ($|\Delta\mathrm{IR}| = 1.59$) nor BoundaryTPS ($1.30$) conditions on class labels during selection, both inherit the source set's class proportions without explicit class-rebalancing machinery.

We call for a discussion about current limitations.

The first is BoundaryTPS's reliance on the nearest non-target neighbor as its boundary signal. BoundaryTPS's performance is competitive only when $k = 1$, with mean G-Mean dropping from $0.759$ at $k = 1$ to $0.523$ at $k = 3$ and $0.431$ at $k = 20$. Additionally, BoundaryTPS is not robust to label noise, falling $0.213$ G-Mean points over the $0\%$--$20\%$ symmetric label-noise range ($0.568 \to 0.355$) and below the baseline at every $\eta > 0\%$. The vertex weight is computed from the nearest non-target neighbors of each target point, and there is no fallback when that signal degrades, whether because additional neighbors stop being informative (the $k$-sensitivity) or because the labels defining "non-target" have been flipped (the noise amplification). This is a scope condition specific to BoundaryTPS, not on topological prototype selection as a whole, since TPS's two-stage construction retains an interior-typical sample independent of boundary weights and produces an increase over the baseline as $\eta$ grows from $+0.011$ at $\eta = 0\%$ to $+0.042$ at $\eta = 20\%$, a pattern shared with explicit noise editors. 

The second is the aggregate classification trade-off. Both methods retain points based on participation in persistence features rather than on local discriminative contribution, and the aggregate mean G-Mean (TPS at $0.79$ and BoundaryTPS at $0.76$) places the proposed methods in the middle of the field, between the top-tier classical methods (SPOTGreedy and KMeans at $0.84$) and the bottom-tier ones (IB3 $0.70$, ICF $0.69$, CNN+ENN $0.53$). Despite the middling aggregate performance, TPS achieves rank-$1$ G-Mean on $11.3\%$ of (dataset, fold, classifier) combinations and rank-$3$-or-better on $25.7\%$, with BoundaryTPS at $9.9\%$ and $18.4\%$. The boundary-proximity principle from the prototype-selection literature predicts that boundary-concentrated selection should help classification, and the current implementations evidently realize that benefit on some classifier–dataset pairings but not others. What "boundary" means for downstream classification performance — whether the same boundary a topological selector identifies is the boundary a given classifier wants prototypes near — remains open. A complexity-segmented evaluation based on dataset characteristics (such as those described in ~\cite{ho_complexity_2002, mollineda_data_2005}) would help locate where each method does and does not improve on the baseline. We leave both to future work.

We advocate for BoundaryTPS as the appropriate choice when topology preservation is the primary criterion. Its weighted filtration achieves significantly better $H_1$ persistence-diagram preservation than every classical compression method evaluated. It is also the appropriate choice when the input data is clean enough that the nearest-non-target neighbor signal is reliable, since both its $k$-sensitivity and its noise amplification trace back to that signal. TPS is the appropriate choice when label noise is a possibility (or class overlap is high), or when broader interior coverage of the class manifold is desirable.

\subsection{Future Work}\label{sec:future_work}

TPS and BoundaryTPS are two specific positions within a broader design space for topological prototype selection that the existing prototype-selection taxonomy does not resolve. The categorization partitions existing methods into condensation, edition, hybrid, competence-based, optimization-based, and clustering-based families. A topological family of selectors organized along its own design axes --- the choice of underlying complex, the weighting scheme, and the feature-selection rule applied to the resulting persistence diagram or summary structure --- is a natural extension for topological-based taxonomy of prototype selectors. The directions below describe extensions across the remaining axes.

%Substituting a witness complex for the Rips complex in the filtration stages would reduce the asymptotic cost of TPS and BoundaryTPS on large datasets without materially degrading the topological signal that drives prototype retention, since the landmark structure already encodes a form of geometric summarization compatible with the goals of prototype selection. 

A natural extension concerns the choice of underlying simplicial complex and its interaction with the input representation. Both methods presented here rely on Rips filtrations, which are attractive for their simplicity and well-understood theoretical properties, but scale poorly with sample size and ambient dimension. Witness complexes offer an alternative by distinguishing a small set of landmark points from a larger set of witnesses, producing filtrations whose size is governed by the number of landmarks rather than the full training set while still recovering the persistent homology of the underlying space under mild sampling conditions~\cite{silva_topological_2004}. A similar idea motivates the study of Flood complexes for even larger datasets where $n$ approaches the millions~\cite{graf_flood_2026}. Other complexes, such as alpha complexes in low-dimensional settings or DTM-based filtrations in the presence of noise~\cite{anai_dtm-based_2018}, present additional axes of variation that may interact differently with the objectives motivating the present framework. 

As mentioned, small perturbations to the underlying dataset produce bounded changes in the persistent diagram \cite{cohen-steiner_stability_2007}. However, any preprocessing step that alters that topology rewrites the persistence diagram and the prototype set with it. Truncated SVD in latent semantic analysis is a known example that can flatten or destroy the topological signal, with severity depending on which features the filtration is built to detect ~\cite{ghafuri_singular-value-decomposition-based_2024}. We advocate caution using TPS and BoundaryTPS for such tasks where topology-destroying preprocessing is used prior to prototype selection.

PH is also metric dependent \cite{grande_non-isotropic_2023}. The filtration radii, vertex weights, and resulting persistence diagrams all change when the underlying distance metric changes, and the empirical evaluation throughout this work used the Euclidean metric. The relative behavior of TPS, BoundaryTPS, and the classical baselines under non-Euclidean metrics is therefore an open question that the present study does not address. As a preliminary probe motivating this direction, Table~\ref{tab:breast_metrics} reports evaluation metrics on the high-dimensional Breast Cancer microarray dataset (OpenML ID: 45085, $n = 97$, $p = 24{,}481$) which is separate from the BreastCancer dataset used in the real data analysis. We calculate prototypes under Manhattan distance, a standard choice for high-dimensional data due to the convergence of distances in Euclidean space with increasing dimensionality ~\cite{aggarwal_surprising_2001}.  BoundaryTPS offers a positive $\Delta$G-Mean of $+1.72$, while TPS completely collapses with a $\Delta$G-Mean of $-14.75$. The magnitude of the gap alone warrants systematic investigation across multiple datasets and non-Euclidean metrics. Such an investigation is the most practically consequential extension of this work for deployment outside the curated-tabular setting on which the main evaluation rests.

\begin{table}[tb]
\centering
\caption{Performance on Breast (OpenML 45085) under Manhattan distance, averaged across six classifiers and 10 outer folds. Sorted by G-Mean. Baseline G-Mean is 0.571.}
\label{tab:breast_metrics}
\small
\begin{tabular}{lrrrrrr}
\hline
\hline
Method & G-Mean & F1 & Acc.\ & $\Delta$G-Mean (pp) & Red.\ \% & Time (s) \\
\hline
KMeans      & 0.619 & 0.631 & 0.642 & $+4.87$ & 72.52 & 0.14 \\
CNN         & 0.616 & 0.631 & 0.644 & $+4.54$ & 36.88 & 0.29 \\
BoundaryTPS & 0.588 & 0.603 & 0.615 & $+1.72$ & 78.21 & 0.12 \\
IB3         & 0.531 & 0.582 & 0.611 & $-3.93$ & 57.28 & 0.11 \\
SPOTGreedy  & 0.531 & 0.586 & 0.618 & $-3.96$ & 70.25 & 0.05 \\
CNN+ENN     & 0.489 & 0.539 & 0.564 & $-8.14$ & 76.40 & 0.41 \\
DROP3       & 0.480 & 0.549 & 0.594 & $-9.04$ & 87.52 & 0.45 \\
ICF         & 0.469 & 0.559 & 0.612 & $-10.19$ & 86.14 & 0.44 \\
TPS         & 0.423 & 0.493 & 0.538 & $-14.75$ & 86.25 & 0.18 \\
\hline
\hline
\end{tabular}
\end{table}

Another possible direction is to consider topological alternatives to persistent homology such as Mapper-based constructions~\cite{singh_topological_2007}. Mapper produces a summary of a dataset by covering the image of a filter function and clustering the preimages, yielding a graph or higher-dimensional complex whose nodes represent regions of the data and whose edges encode overlap. Applied class-by-class and projected onto the original point cloud, a Mapper graph would identify structurally distinct regions within each class, and synthetic prototypes could be generated from such nodes (projected back onto the original data cloud) or selected according to centrality, branch membership, or position relative to the between-class boundary. The principal obstacle is Mapper's well-known sensitivity to its filter function, cover resolution, and clustering hyperparameters, though recent work on more stable variants has begun to address these limitations~\cite{chalapathi_adaptive_2021}.

Lastly, we discuss combining topology-informed selection with existing geometric prototype selection methods rather than treat them as alternatives. Methods such as DROP3, ICF, and CNN operate on local neighborhood relations and decision-boundary heuristics with no mechanism for incorporating global structural information about a class. One way to bridge this gap would be to compute persistence-based descriptors and use them to augment the feature representation that geometric selectors operate on. A geometric selector would then make retention decisions in a feature space enriched by topological context rather than in the raw input space alone. A second avenue is hybrid scoring, in which a prototype's retention score is a function of both its geometric utility (e.g., contribution to a nearest-neighbor decision boundary) and its topological role (e.g., participation in long-lived merge events or location in a Mapper node of high boundary salience). Hybrid selectors of this form would, in principle, capture both the local discriminative information that geometric methods resolve and the global organizational structure that topological methods detect, two signals neither family can recover alone. 

Pursuing these directions, individually or in combination, would do more than introduce additional methods. Each extension systematically explores a different axis of the design space sketched at the opening of this section. As more realizations are characterized, the structural properties documented in this work (topology preservation, stability, IR behavior, scalable cost, classification performance) can be evaluated against the design choices that produce them, and patterns of specialization across realizations would become visible. 

As a result we do not attempt to fully develop the topological taxonomy in detail here, but rather we identify the research direction that would do so, and we expect the development of topology-based selectors to drive its characterization over time as more entries fill out the design space.

\newpage

\section{Appendix}

\subsection{Algorithm Pseudocode}
\begin{algorithm}[h]
\caption{Boundary-Conscious Topological Prototype Selection (BoundaryTPS)}
\label{alg:btps}
\begin{algorithmic}[1]
\Require Feature matrix $\mathcal X \in \mathbb{R}^{n \times d}$, labels $y \in \mathcal Y =\{1,\dots,C\}^n$, neighbors $k$, quantile $q$, tolerance $\tau$, homology dimension $p$
\Ensure Prototype indices $S \subseteq \{1, \dots, n\}$
\State $S \leftarrow \emptyset$
\For{each class $c \in \mathcal{Y}$}
    \State $T_c \leftarrow \{i : y_i = c\}$, \quad $N_c \leftarrow \{i : y_i \neq c\}$
    \For{each $x_i$, $i \in T_c$}
        \State Find $k$ nearest neighbors of $x_i$ in $\{x_j : j \in N_c\}$
        \State $w_i \leftarrow \sum_{j \in \mathrm{kNN}_{\neg c}(x_i)} d(x_i, x_j)$
    \EndFor
    \State $D \leftarrow$ pairwise distance matrix of $\{x_i : i \in T_c\}$
    \State $\tilde{D}_{ij} \leftarrow \max(D_{ij}, w_i, w_j)$ for $i \neq j$; \quad $\tilde{D}_{ii} \leftarrow w_i$
    \State Build Vietoris-Rips filtration on $\tilde{D}$
    \State Compute $H_p$ persistence; extract features $F = \{(b_i, d_i, \ell_i)\}$
    \State $\ell_q \leftarrow \mathrm{QuantInt}(q, \{\ell_i\})$
    \If{$\tau = 0$}
        \State $F_{\mathrm{sel}} \leftarrow \{f \in F : \ell_f \text{ is closest to } \ell_q\}$
    \Else
        \State $F_{\mathrm{sel}} \leftarrow \{f \in F : \ell_q(1-\tau) \leq \ell_f \leq \ell_q(1+\tau)\}$
    \EndIf
    \If{$p = 0$}
        \State Union-Find up to $\max\{d_f : f \in F_{\mathrm{sel}}\}$
        \State $V_c \leftarrow$ all vertices in components involved in any merge event
    \Else
        \State $V_c \leftarrow$ all vertices of $(p+1)$-simplices added up to $\max\{d_f : f \in F_{\mathrm{sel}}\}$
    \EndIf
    \State $S \leftarrow S \cup V_c$
\EndFor
\State \Return $S$
\end{algorithmic}
\end{algorithm}

\begin{algorithm}[h]
\caption{Topological Prototype Selection (TPS)}
\label{alg:tps}
\begin{algorithmic}[1]
\Require Feature matrix $\mathcal X \in \mathbb{R}^{n \times d}$, labels $y \in \mathcal Y = \{1,\dots,C\}^n$, neighbors $k$, quantile $q$, homology dimension $p$
\Ensure Prototype indices $S \subseteq \{1, \dots, n\}$
\State $S \leftarrow \emptyset$
\For{each class $c \in \mathcal{Y}$}
    \State $T_c \leftarrow \{i : y_i = c\}$, \quad $N_c \leftarrow \{i : y_i \neq c\}$
    \State \Comment{\textbf{Stage 1: Boundary-Neighborhood Filtration}}
    \For{each $x_i$, $i \in T_c$}
        \State Find $k$ nearest neighbors of $x_i$ in $\{x_j : j \in N_c\}$
    \EndFor
    \State $B_c \leftarrow T_c \cup \{\text{unique nontarget neighbors}\}$
    \State $D \leftarrow$ pairwise distance matrix of $\{x_i : i \in B_c\}$
    \State Build Vietoris-Rips filtration on $D$
    \State Compute $H_p$ persistence; extract features $F = \{(b_i, d_i, \ell_i)\}$
    \State $\ell_q \leftarrow \mathrm{QuantInt}(q, \{\ell_i\})$
    \State $F_{\mathrm{sel}} \leftarrow \{f \in F : \ell_f \text{ is closest to } \ell_q\}$
    \If{$p = 0$}
        \State Union-Find up to $\max\{d_f : f \in F_{\mathrm{sel}}\}$
        \State $V_1 \leftarrow$ all vertices in components involved in any merge event
    \Else
        \State $V_1 \leftarrow$ all vertices of $(p+1)$-simplices up to $\max\{d_f : f \in F_{\mathrm{sel}}\}$
    \EndIf
    \State $V_1 \leftarrow V_1 \cap T_c$ \Comment{Retain only target-class points}
    \State \Comment{\textbf{Stage 2: Interior Filtration}}
    \State $D' \leftarrow$ pairwise distance matrix of $\{x_i : i \in V_1\}$
    \State Build Vietoris-Rips filtration on $D'$
    \State Compute $H_p$ persistence; extract features $F' = \{(b_i, d_i, \ell_i)\}$
    \State $\ell_{\mathrm{mean}} \leftarrow \frac{1}{|F'|} \sum_{f_i \in F'} \ell_i$
    \State $F'_{\mathrm{sel}} \leftarrow \{f \in F' : \ell_f \text{ is closest to } \ell_{\mathrm{mean}}\}$
    \If{$p = 0$}
        \State Union-Find up to $\max\{d_f : f \in F'_{\mathrm{sel}}\}$
        \State $V_2 \leftarrow$ all vertices in components involved in merge events
    \Else
        \State $V_2 \leftarrow$ vertices of $(p+1)$-simplices up to $\max\{d_f : f \in F'_{\mathrm{sel}}\}$
    \EndIf
    \State $S \leftarrow S \cup V_2$
\EndFor
\State \Return $S$
\end{algorithmic}
\end{algorithm}

\clearpage

\subsection{Simulated Datasets Selected Prototypes}

\begin{figure}[h!]
    \centering
    \includegraphics[width=.53\linewidth]{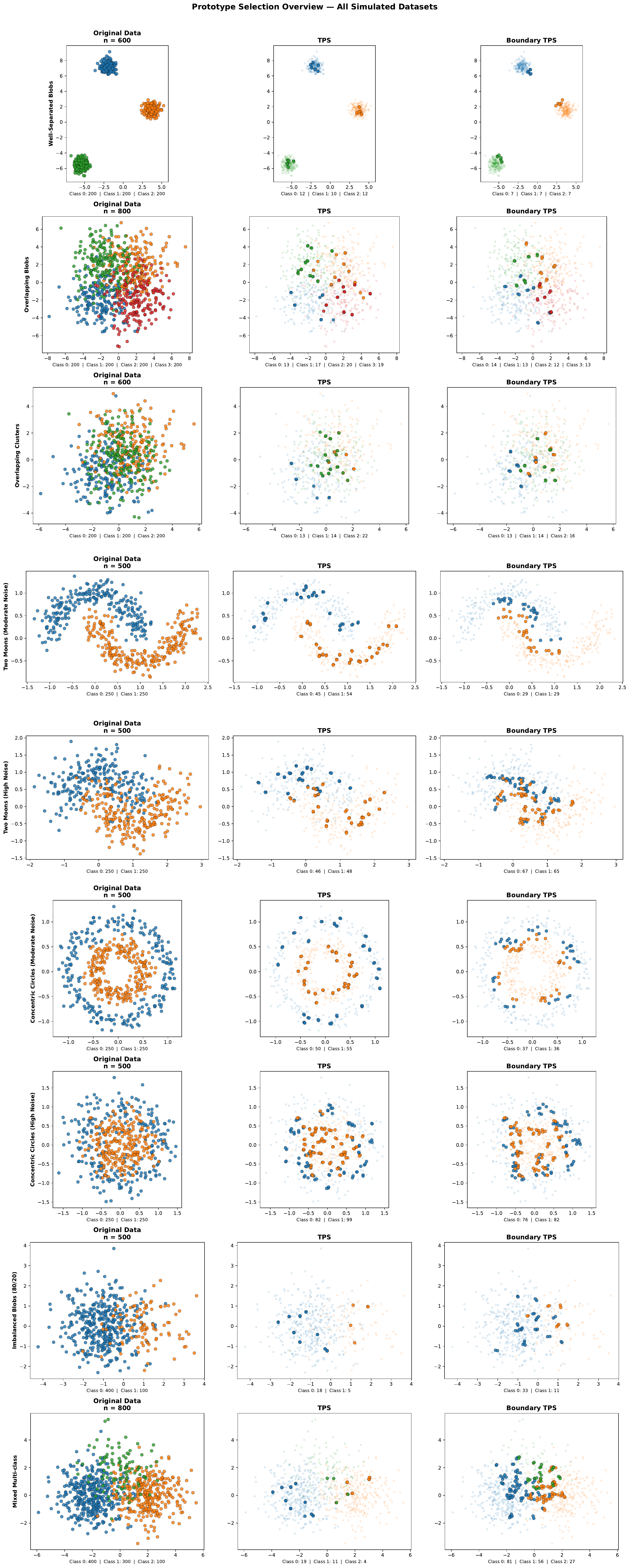}
\end{figure}

\clearpage

\subsection{Simulated Dataset Results}

\begin{table}[h]
\centering
\small
\caption{TPS and BoundaryTPS performance on simulated datasets under stratified 10-fold cross-validation. G-Mean values are reported as mean $\pm$ standard deviation across folds, with $\Delta$G-Mean (percentage point change relative to the baseline) shown below each entry.}
\label{tab:simulated_results}
\resizebox{\textwidth}{!}{%
\begin{tabular}{llcccccccc}
\hline\hline
 & & \multicolumn{6}{c}{\textbf{G-Mean $\pm$ Std}} & & \\
 & & \multicolumn{6}{c}{($\Delta$\textbf{G-Mean})} & & \\
\cline{3-8}
\textbf{Dataset} & \textbf{Variant} & \textbf{1-NN} & \textbf{3-NN} & \textbf{5-NN} & \textbf{Lin.\ SVM} & \textbf{RF} & \textbf{MLP} & \textbf{Red.\ (\%)} & \textbf{Time (s)} \\
\hline
Well-Sep.\ Blobs & TPS & $1.00 \pm .00$ & $1.00 \pm .00$ & $1.00 \pm .00$ & $1.00 \pm .00$ & $1.00 \pm .00$ & $1.00 \pm .00$ & 93.7 & 0.03 \\
 & & $(+0.0)$ & $(+0.0)$ & $(+0.0)$ & $(+0.0)$ & $(+0.0)$ & $(+0.0)$ & & \\
 & BoundaryTPS & $1.00 \pm .00$ & $1.00 \pm .00$ & $1.00 \pm .00$ & $1.00 \pm .00$ & $1.00 \pm .00$ & $1.00 \pm .00$ & 96.7 & 0.02 \\
 & & $(+0.0)$ & $(+0.0)$ & $(+0.0)$ & $(+0.0)$ & $(+0.0)$ & $(+0.0)$ & & \\[3pt]
Overlap.\ Blobs & TPS & $0.63 \pm .08$ & $0.66 \pm .08$ & $0.65 \pm .08$ & $0.71 \pm .06$ & $0.68 \pm .07$ & $0.70 \pm .05$ & 85.7 & 0.11 \\
 & & $(-0.2)$ & $(-1.4)$ & $(-5.9)$ & $(-2.0)$ & $(-2.0)$ & $(-2.8)$ & & \\
 & BoundaryTPS & $0.61 \pm .08$ & $0.61 \pm .08$ & $0.69 \pm .06$ & $0.69 \pm .07$ & $0.67 \pm .06$ & $0.68 \pm .07$ & 80.5 & 0.03 \\
 & & $(-2.1)$ & $(-6.6)$ & $(-1.7)$ & $(-3.7)$ & $(-3.0)$ & $(-4.9)$ & & \\[3pt]
Overlap.\ Clusters & TPS & $0.46 \pm .05$ & $0.45 \pm .08$ & $0.50 \pm .08$ & $0.52 \pm .08$ & $0.43 \pm .08$ & $0.47 \pm .09$ & 81.8 & 0.10 \\
 & & $(-1.6)$ & $(-2.4)$ & $(+0.5)$ & $(-2.4)$ & $(-2.4)$ & $(-5.4)$ & & \\
 & BoundaryTPS & $0.51 \pm .07$ & $0.50 \pm .06$ & $0.48 \pm .07$ & $0.46 \pm .11$ & $0.44 \pm .10$ & $0.48 \pm .08$ & 85.6 & 0.02 \\
 & & $(+3.0)$ & $(+2.4)$ & $(-2.2)$ & $(-7.7)$ & $(-1.2)$ & $(-4.3)$ & & \\[3pt]
Two Moons (Mod.) & TPS & $0.98 \pm .03$ & $0.98 \pm .03$ & $0.95 \pm .05$ & $0.85 \pm .06$ & $0.90 \pm .09$ & $0.86 \pm .06$ & 84.9 & 0.04 \\
 & & $(-2.1)$ & $(-1.5)$ & $(-4.6)$ & $(-0.8)$ & $(-9.2)$ & $(-12.0)$ & & \\
 & BoundaryTPS & $0.88 \pm .09$ & $0.93 \pm .05$ & $0.90 \pm .06$ & $0.76 \pm .10$ & $0.87 \pm .13$ & $0.75 \pm .13$ & 85.0 & 0.02 \\
 & & $(-11.5)$ & $(-6.3)$ & $(-9.3)$ & $(-9.8)$ & $(-11.7)$ & $(-23.0)$ & & \\[3pt]
Two Moons (High) & TPS & $0.82 \pm .06$ & $0.85 \pm .06$ & $0.83 \pm .05$ & $0.80 \pm .06$ & $0.83 \pm .06$ & $0.82 \pm .06$ & 85.4 & 0.05 \\
 & & $(+0.8)$ & $(-0.7)$ & $(-3.8)$ & $(-3.7)$ & $(-3.7)$ & $(-4.4)$ & & \\
 & BoundaryTPS & $0.80 \pm .07$ & $0.80 \pm .09$ & $0.82 \pm .10$ & $0.78 \pm .08$ & $0.83 \pm .04$ & $0.77 \pm .09$ & 81.1 & 0.02 \\
 & & $(-2.1)$ & $(-5.8)$ & $(-4.9)$ & $(-5.5)$ & $(-3.5)$ & $(-9.8)$ & & \\[3pt]
Circles (Mod.) & TPS & $0.93 \pm .06$ & $0.92 \pm .06$ & $0.94 \pm .04$ & $0.20 \pm .27$ & $0.91 \pm .06$ & $0.93 \pm .08$ & 83.0 & 0.04 \\
 & & $(-5.4)$ & $(-6.1)$ & $(-4.8)$ & $(-28.1)$ & $(-6.2)$ & $(-5.5)$ & & \\
 & BoundaryTPS & $0.90 \pm .11$ & $0.95 \pm .04$ & $0.92 \pm .05$ & $0.35 \pm .24$ & $0.94 \pm .05$ & $0.90 \pm .10$ & 86.0 & 0.02 \\
 & & $(-8.6)$ & $(-3.9)$ & $(-6.8)$ & $(-13.6)$ & $(-3.1)$ & $(-8.2)$ & & \\[3pt]
Circles (High) & TPS & $0.70 \pm .08$ & $0.68 \pm .10$ & $0.72 \pm .08$ & $0.32 \pm .23$ & $0.74 \pm .03$ & $0.75 \pm .07$ & 81.7 & 0.05 \\
 & & $(-4.0)$ & $(-9.8)$ & $(-7.0)$ & $(-20.0)$ & $(-3.7)$ & $(-6.3)$ & & \\
 & BoundaryTPS & $0.69 \pm .05$ & $0.72 \pm .04$ & $0.74 \pm .05$ & $0.48 \pm .09$ & $0.72 \pm .07$ & $0.77 \pm .07$ & 78.6 & 0.02 \\
 & & $(-4.4)$ & $(-6.5)$ & $(-4.5)$ & $(-4.1)$ & $(-6.0)$ & $(-4.5)$ & & \\[3pt]
Imbal.\ (80/20) & TPS & $0.83 \pm .09$ & $0.80 \pm .11$ & $0.80 \pm .07$ & $0.83 \pm .10$ & $0.82 \pm .08$ & $0.79 \pm .08$ & 89.7 & 0.04 \\
 & & $(+12.4)$ & $(+9.8)$ & $(+7.5)$ & $(+2.6)$ & $(+8.2)$ & $(+1.5)$ & & \\
 & BoundaryTPS & $0.69 \pm .11$ & $0.71 \pm .09$ & $0.75 \pm .12$ & $0.79 \pm .10$ & $0.62 \pm .14$ & $0.70 \pm .13$ & 87.2 & 0.02 \\
 & & $(-1.4)$ & $(+0.4)$ & $(+2.5)$ & $(-1.8)$ & $(-11.2)$ & $(-7.9)$ & & \\[3pt]
Mixed Multiclass & TPS & $0.69 \pm .10$ & $0.70 \pm .09$ & $0.70 \pm .08$ & $0.76 \pm .06$ & $0.66 \pm .11$ & $0.73 \pm .08$ & 88.7 & 0.07 \\
 & & $(+2.3)$ & $(+1.4)$ & $(+1.0)$ & $(+1.3)$ & $(-3.4)$ & $(-3.4)$ & & \\
 & BoundaryTPS & $0.60 \pm .22$ & $0.64 \pm .10$ & $0.58 \pm .21$ & $0.73 \pm .08$ & $0.65 \pm .07$ & $0.70 \pm .12$ & 86.2 & 0.03 \\
 & & $(-6.0)$ & $(-5.4)$ & $(-11.1)$ & $(-1.8)$ & $(-4.3)$ & $(-6.1)$ & & \\
\hline\hline
\end{tabular}
}
\end{table}

\newpage
\clearpage

\subsection{Per Dataset Friedman Ranks for $H_1$ Homology on Wasserstein-1 Distance}
\begin{table}[h]
\centering
\small
\begin{tabular}{lccccccccc}
\hline\hline
\textbf{Dataset} & \textbf{BoundaryTPS} & \textbf{CNN} & \textbf{TPS} & \textbf{IB3} & \textbf{CNN+ENN} & \textbf{DROP3} & \textbf{KMeans} & \textbf{SPOTGreedy} & \textbf{ICF} \\
\hline
BloodTransfusion & 5.0 & \textbf{1.0} & 3.0 & 2.0 & 4.0 & 6.0 & 9.0 & 7.0 & 8.0 \\
BreastCancer & 2.0 & \textbf{1.0} & 6.0 & 4.0 & 3.0 & 5.0 & 7.0 & 8.0 & 9.0 \\
CardiotocographyA & \textbf{1.0} & 7.0 & 2.0 & 8.0 & 9.0 & 6.0 & 4.0 & 5.0 & 3.0 \\
Diabetes & 3.0 & \textbf{1.0} & 6.0 & 2.0 & 4.0 & 5.0 & 7.0 & 8.0 & 9.0 \\
Digits & \textbf{1.0} & 3.0 & 2.0 & 4.0 & 8.0 & 7.0 & 6.0 & 5.0 & 9.0 \\
Ionosphere & \textbf{1.0} & 5.0 & 2.0 & 6.0 & 3.0 & 4.0 & 8.0 & 7.0 & 9.0 \\
Iris & 2.0 & 7.0 & 4.0 & 6.0 & 3.0 & 5.0 & 9.0 & 8.0 & \textbf{1.0} \\
KCvsKP & 2.0 & 3.0 & \textbf{1.0} & 5.0 & 6.0 & 4.0 & 9.0 & 8.0 & 7.0 \\
MiceProtein & \textbf{1.0} & 5.0 & 2.0 & 7.0 & 8.0 & 9.0 & 3.0 & 4.0 & 6.0 \\
Ozone & 3.0 & 2.0 & 6.0 & \textbf{1.0} & 4.0 & 5.0 & 8.0 & 9.0 & 7.0 \\
Satimage & 2.0 & \textbf{1.0} & 3.0 & 4.0 & 7.0 & 6.0 & 8.0 & 9.0 & 5.0 \\
Sonar & \textbf{1.0} & 2.0 & 4.0 & 3.0 & 7.0 & 5.0 & 6.0 & 9.0 & 8.0 \\
Spambase & 2.0 & \textbf{1.0} & 5.0 & 3.0 & 4.0 & 6.0 & 8.0 & 7.0 & 9.0 \\
Wilt & 3.0 & 2.0 & 4.0 & \textbf{1.0} & 5.0 & 6.0 & 9.0 & 8.0 & 7.0 \\
Wine & \textbf{1.0} & 7.0 & 4.0 & 8.0 & 6.0 & 5.0 & 2.0 & 3.0 & 9.0 \\
\hline
Avg. Rank & \textbf{2.00} & 3.20 & 3.60 & 4.27 & 5.40 & 5.60 & 6.87 & 7.00 & 7.07 \\
\hline\hline
\\
\end{tabular}
\caption{Per-dataset Friedman ranks for the H$_1$ homology on raw Wasserstein$_1$ distance. Lower rank indicates a more faithful preservation of the persistence diagram. Bolded values are per-dataset winner. The bottom row is the mean rank across datasets.}
\label{tab:real_topology_per_dataset_rank}
\end{table}

\newpage
\clearpage

\subsection{Betti and Euler Curve Diagrams}

\begin{figure}[h!]
    \centering
    \includegraphics[width=.9\linewidth]{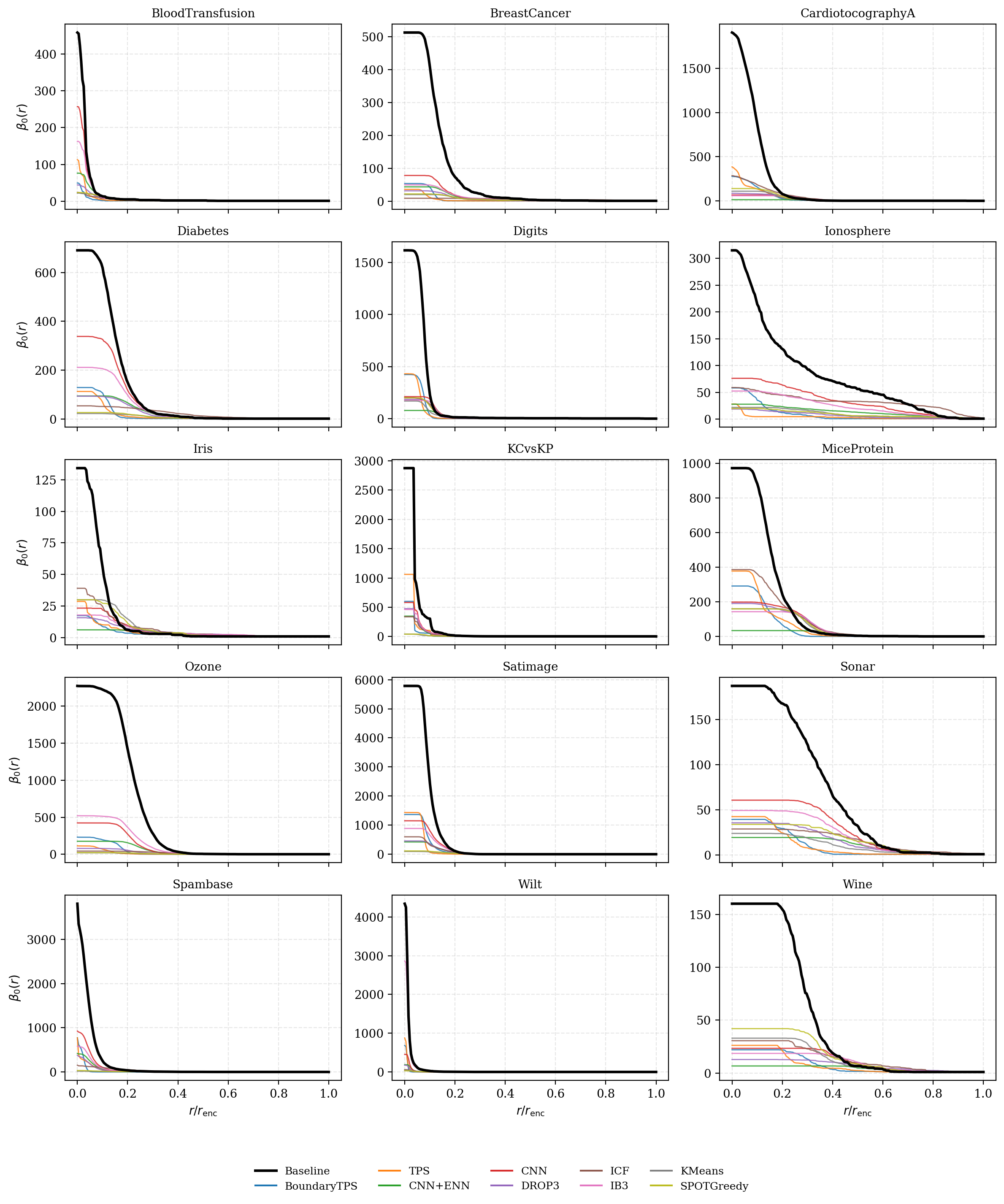}
    \caption{Zero-dimensional Betti curves $\beta_0(r)$ for the full training set
    $X_{\mathrm{train}}$ (thick black) and for each prototype-selection method
    (colored), averaged across the 10 stratified CV folds, across the 15 real
    datasets. The horizontal axis is the filtration radius normalized by
    $r_{\mathrm{enc}}(X_{\mathrm{train}})$. Because every prototype set has
    cardinality $|S| < |X_{\mathrm{train}}|$, $\beta_0(0^+) \leq |S|$ by
    construction, so prototype curves necessarily lie below the baseline at
    small~$r$; the diagnostic comparison is the \emph{shape} of the merging
    cascade not the absolute vertical offset.}
    \label{fig:ho_betti}
\end{figure}

\clearpage

\begin{figure}[h!]
    \centering
    \includegraphics[width=\linewidth]{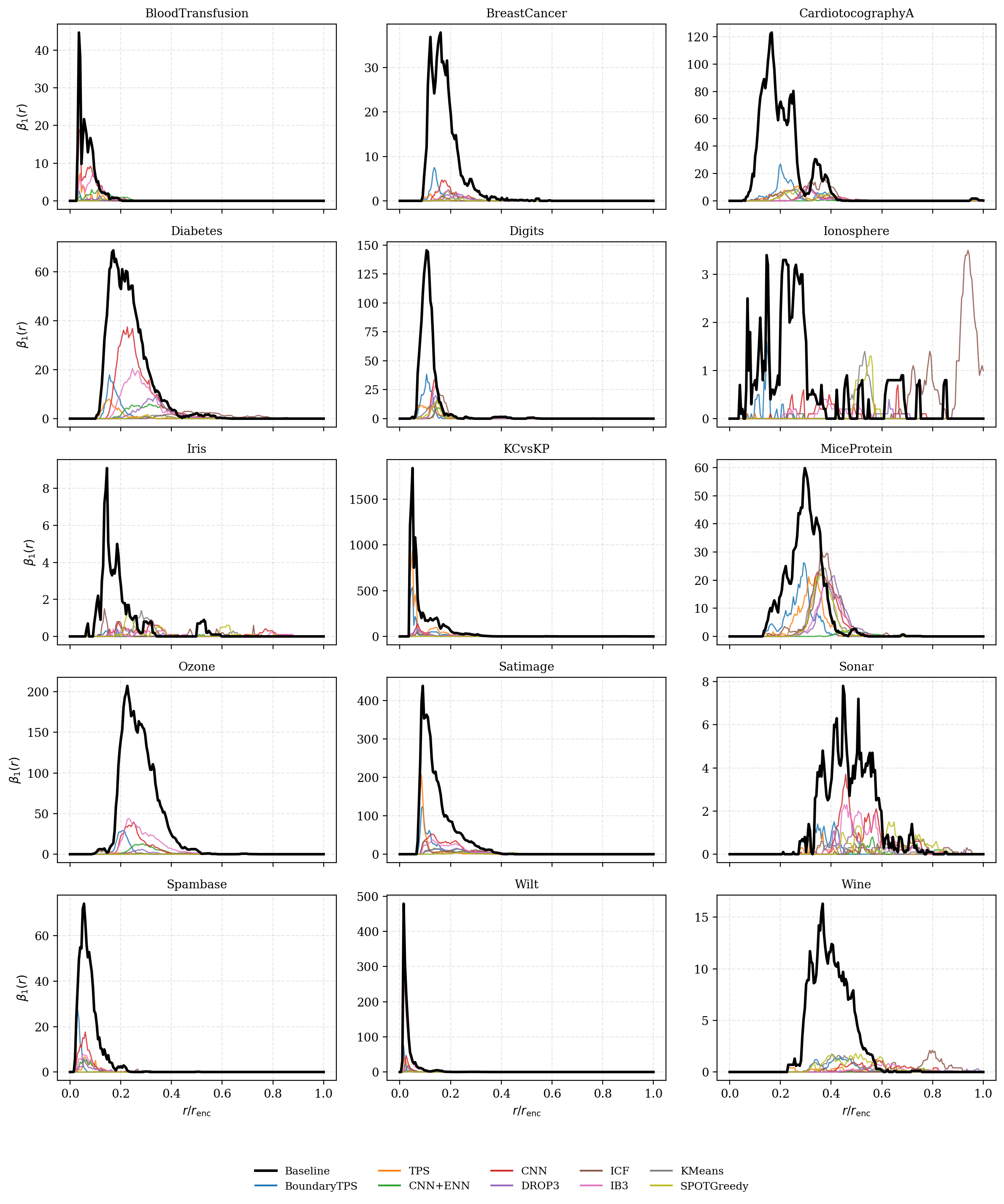}
    \caption{First-dimensional Betti curves $\beta_1(r)$ for the full training set
    $X_{\mathrm{train}}$ (thick black) and for each prototype-selection method
    (colored), averaged across the 10 stratified CV folds, across the 15 real
    datasets. The horizontal axis is the filtration radius normalized by
    $r_{\mathrm{enc}}(X_{\mathrm{train}})$. Because every prototype set has
    cardinality $|S| < |X_{\mathrm{train}}|$, $\beta_0(0^+) \leq |S|$ by
    construction, so prototype curves necessarily lie below the baseline at
    small~$r$; the diagnostic comparison is the \emph{shape} of the merging
    cascade not the absolute vertical offset.}
    \label{fig:h1_betti}
\end{figure}

\begin{figure}[h!]
    \centering
    \includegraphics[width=\linewidth]{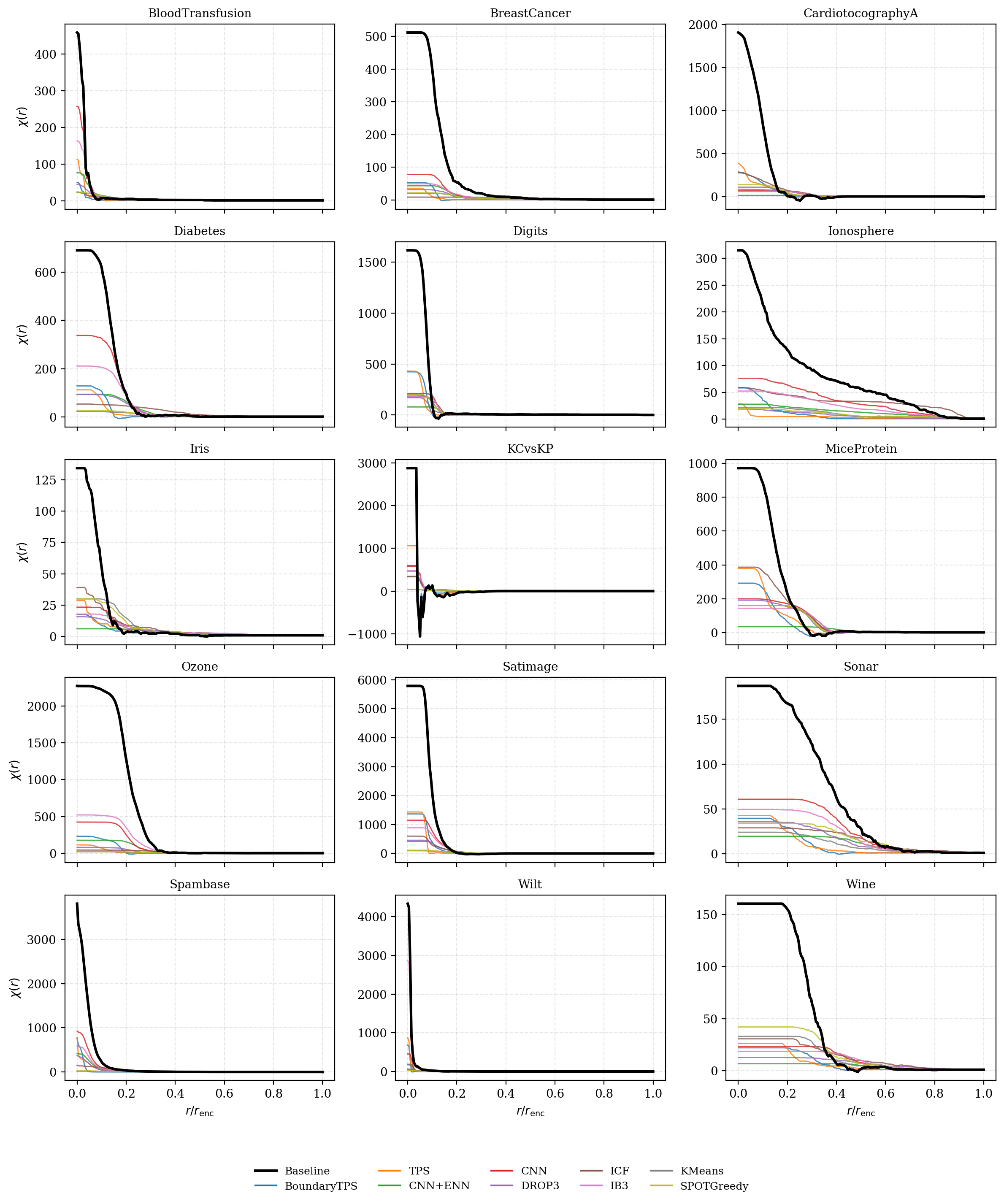}
    \caption{Euler characteristic curves $\chi(r) = \beta_0(r) - \beta_1(r)$ across the 15 real datasets. The thick black curve is the baseline computed on the full training set $X_{\mathrm{train}}$; colored curves are the per-method
    prototype-set curves, averaged across the 10 stratified CV folds. The
    horizontal axis is the filtration radius normalized by the enclosing radius $r_{\mathrm{enc}}(X_{\mathrm{train}})$. The horizontal axis is the filtration radius normalized by
    $r_{\mathrm{enc}}(X_{\mathrm{train}})$. Because every prototype set has
    cardinality $|S| < |X_{\mathrm{train}}|$, $\beta_0(0^+) \leq |S|$ by
    construction, so prototype curves necessarily lie below the baseline at
    small~$r$; the diagnostic comparison is the \emph{shape} of the merging
    cascade not the absolute vertical offset.}
    \label{fig:euler}
\end{figure}
\clearpage

\subsection{Critical Difference Diagrams}

\begin{figure}[h!]
    \centering
    \includegraphics[width=\linewidth]{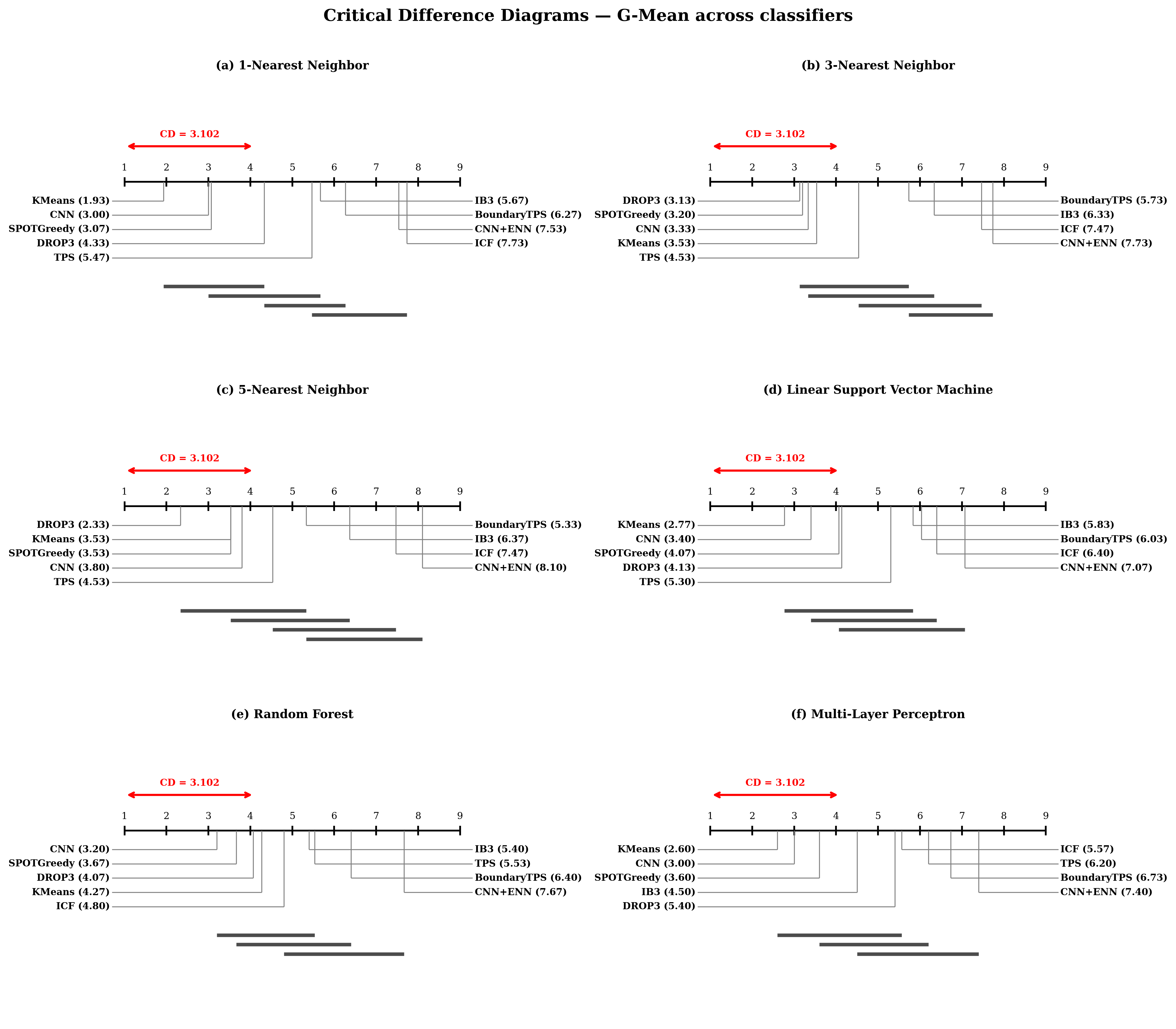}
\end{figure}

\newpage
\clearpage

\subsection{Bonferroni-Dunn Average Ranks with TPS / BoundaryTPS as Control}
\begin{table}[h]
\centering
\small
\begin{tabular}{lcccccc}
\hline\hline
\textbf{Method} & \textbf{1-NN} & \textbf{3-NN} & \textbf{5-NN} & \textbf{Lin.\ SVM} & \textbf{MLP} & \textbf{RF} \\
\hline
KMeans & 1.93$^{\dagger}$ & 3.53 & 3.53 & 2.77 & 2.60$^{\dagger}$ & 4.27 \\
CNN & 3.00 & 3.33 & 3.80 & 3.40 & 3.00$^{\dagger}$ & 3.20 \\
SPOTGreedy & 3.07 & 3.20 & 3.53 & 4.07 & 3.60 & 3.67 \\
DROP3 & 4.33 & 3.13 & 2.33 & 4.13 & 5.40 & 4.07 \\
TPS (focal) & 5.47 & 4.53 & 4.53 & 5.30 & 6.20 & 5.53 \\
IB3 & 5.67 & 6.33 & 6.37 & 5.83 & 4.50 & 5.40 \\
BoundaryTPS & 6.27 & 5.73 & 5.33 & 6.03 & 6.73 & 6.40 \\
ICF & 7.73 & 7.47$^{\dagger}$ & 7.47$^{\dagger}$ & 6.40 & 5.57 & 4.80 \\
CNN+ENN & 7.53 & 7.73$^{\dagger}$ & 8.10$^{\dagger}$ & 7.07 & 7.40 & 7.67 \\
\hline\hline
\\
\end{tabular}
\caption{Bonferroni-Dunn average ranks per classifier on the real-world datasets, with \textbf{TPS} as the control method ($\alpha = 0.05$). Within each classifier, methods are ranked across datasets by mean G-Mean across folds (lower rank means better performance). Cells marked with $^{\dagger}$ differ from the control beyond the Bonferroni-corrected critical difference; $\text{CD}_{\text{BD}} = 2.73$. Row order is by mean rank across classifiers (best at top).}
\label{tab:real_bonferroni_dunn_tps}
\end{table}

\begin{table}[h]
\centering
\small
\begin{tabular}{lcccccc}
\hline\hline
\textbf{Method} & \textbf{1-NN} & \textbf{3-NN} & \textbf{5-NN} & \textbf{Lin.\ SVM} & \textbf{MLP} & \textbf{RF} \\
\hline
KMeans & 1.93$^{\dagger}$ & 3.53 & 3.53 & 2.77$^{\dagger}$ & 2.60$^{\dagger}$ & 4.27 \\
CNN & 3.00$^{\dagger}$ & 3.33 & 3.80 & 3.40 & 3.00$^{\dagger}$ & 3.20$^{\dagger}$ \\
SPOTGreedy & 3.07$^{\dagger}$ & 3.20 & 3.53 & 4.07 & 3.60$^{\dagger}$ & 3.67 \\
DROP3 & 4.33 & 3.13 & 2.33$^{\dagger}$ & 4.13 & 5.40 & 4.07 \\
TPS & 5.47 & 4.53 & 4.53 & 5.30 & 6.20 & 5.53 \\
IB3 & 5.67 & 6.33 & 6.37 & 5.83 & 4.50 & 5.40 \\
BoundaryTPS (focal) & 6.27 & 5.73 & 5.33 & 6.03 & 6.73 & 6.40 \\
ICF & 7.73 & 7.47 & 7.47 & 6.40 & 5.57 & 4.80 \\
CNN+ENN & 7.53 & 7.73 & 8.10$^{\dagger}$ & 7.07 & 7.40 & 7.67 \\
\hline\hline
\\
\end{tabular}
\caption{Bonferroni-Dunn average ranks per classifier on the real-world datasets, with \textbf{BoundaryTPS} as the control method ($\alpha = 0.05$). Within each classifier, methods are ranked across datasets by mean G-Mean across folds (lower rank means better performance). Cells marked with $^{\dagger}$ differ from the control beyond the Bonferroni-corrected critical difference; $\text{CD}_{\text{BD}} = 2.73$. Row order is by mean rank across classifiers (best at top).}
\label{tab:real_bonferroni_dunn_boundarytps}
\end{table}

\subsection{Imbalance Ratio Preservation}
\begin{table}[h]
\centering
\small
\begin{tabular}{lcccc}
\hline\hline
\textbf{Method} & \textbf{Mean} $|\Delta\mathrm{IR}|$ & \textbf{Std} & \textbf{Median} & \textbf{N} \\
\hline
BoundaryTPS & 1.303 & 2.312 & 0.175 & 150 \\
TPS & 1.586 & 2.285 & 0.734 & 150 \\
DROP3 & 2.871 & 4.304 & 0.932 & 150 \\
CNN+ENN & 3.087 & 3.410 & 1.349 & 150 \\
KMeans & 3.202 & 5.305 & 0.681 & 150 \\
SPOTGreedy & 3.202 & 5.305 & 0.681 & 150 \\
CNN & 3.282 & 4.392 & 1.264 & 150 \\
ICF & 3.606 & 5.108 & 0.843 & 150 \\
IB3 & 16.042 & 52.245 & 0.940 & 150 \\
\hline\hline
\\
\end{tabular}
\caption{Per-method absolute imbalance-ratio deviation $|\mathrm{IR}_{\mathrm{proto}} - \mathrm{IR}_{\mathrm{original}}|$ between each method's prototype set and its source training set, aggregated across 150 fold-level observations. Lower value indicates a better preservation of the original class balance.}
\label{tab:real_ir_deviation}
\end{table}

\newpage
\clearpage

\subsection{Hyperparameter Sensitivity}

\begin{figure}[h]
    \centering
    \includegraphics[width=.75\linewidth]{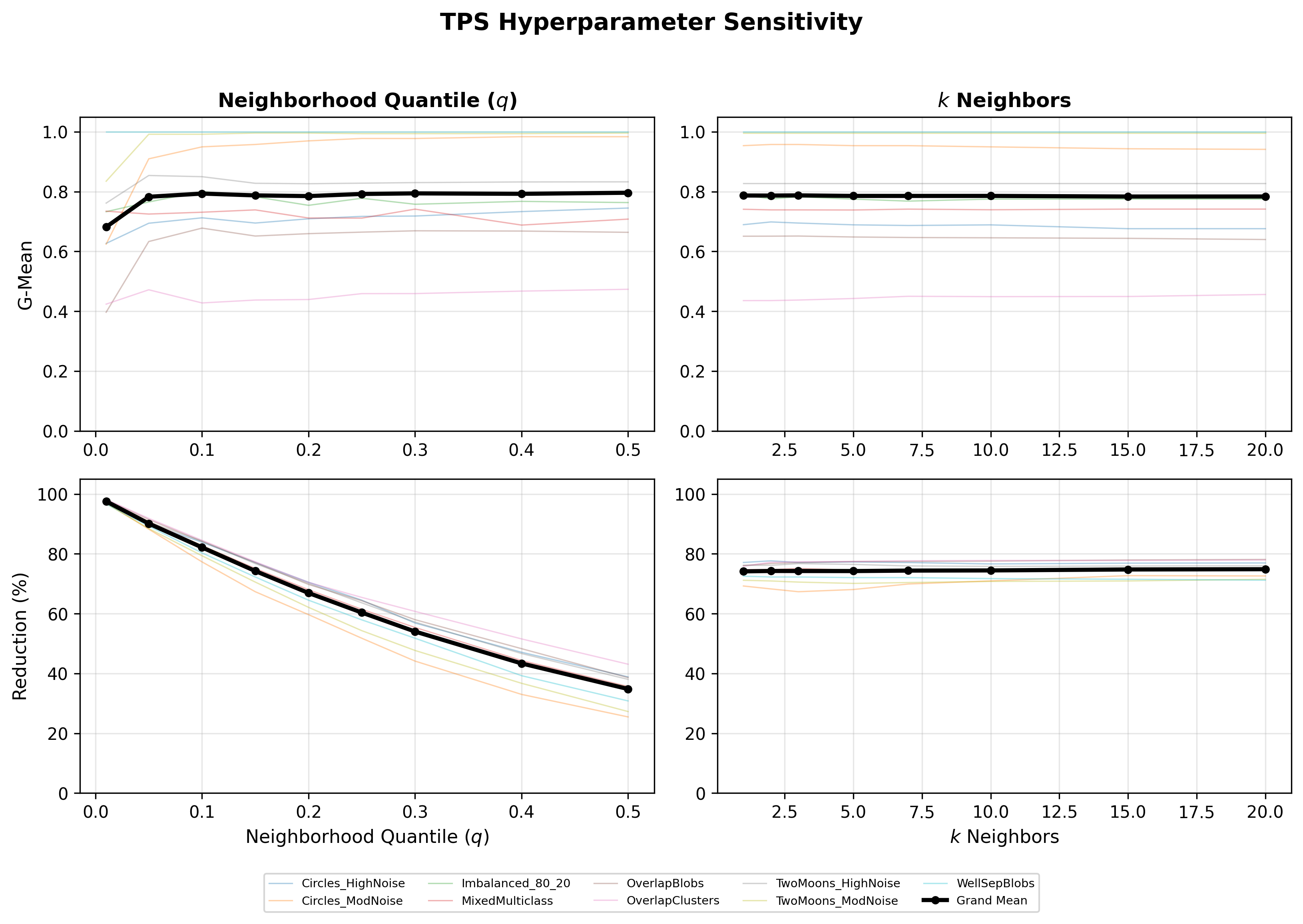}
    \caption{TPS hyperparameter sensitivity: effect of $q$ (left) and $k$ (right) on G-Mean and reduction percentage, averaged across the nine simulated datasets under 1-NN.}
    \label{fig:hp_tps}
\end{figure}

\begin{figure}[h]
    \centering
    \includegraphics[width=\linewidth]{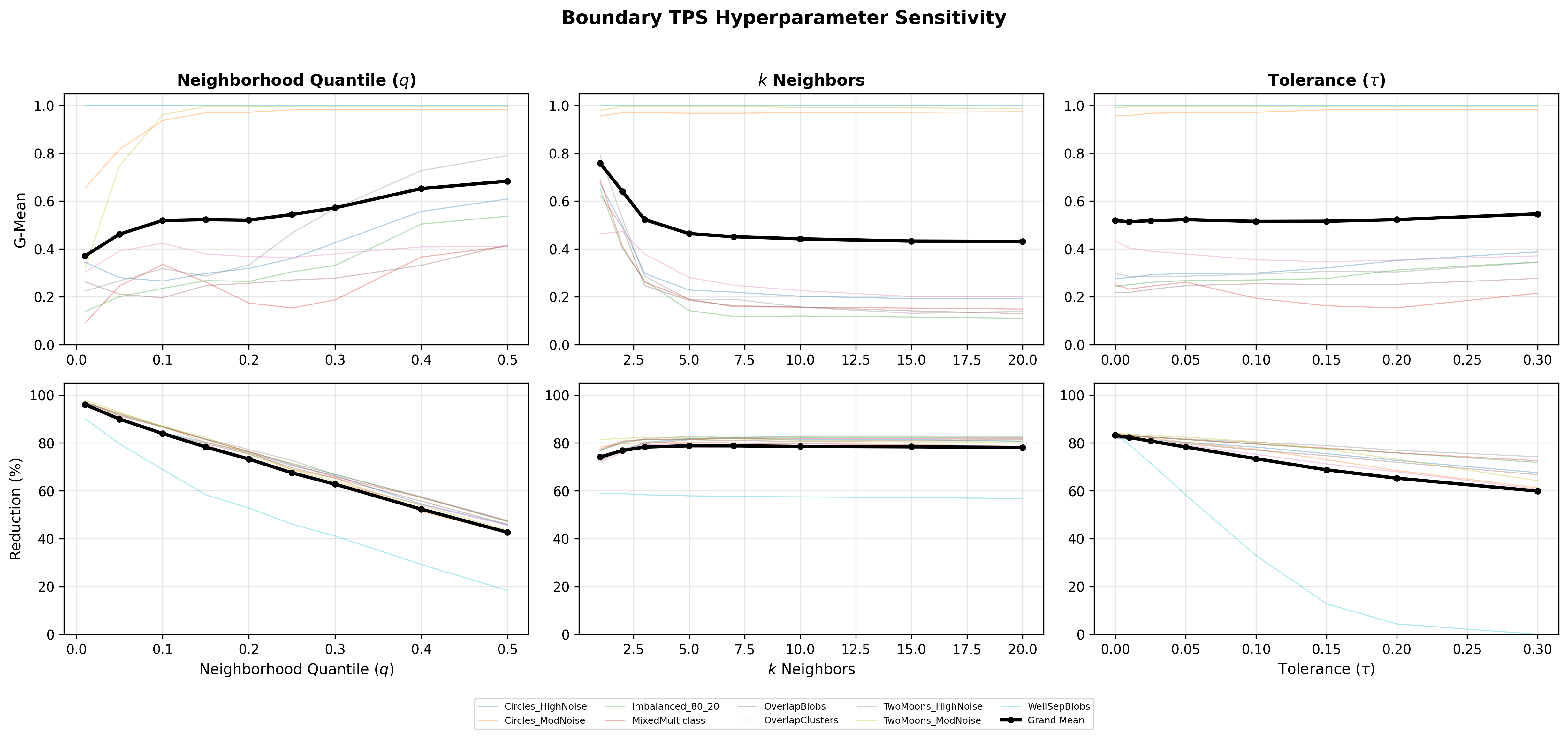}
    \caption{BoundaryTPS hyperparameter sensitivity: effect of $q$ (left), $k$ (center), and $\tau$ (right) on G-Mean and reduction percentage, averaged across the nine simulated datasets under 1-NN.}
    \label{fig:hp_btps}
\end{figure}
\newpage
\clearpage

\section{References}

\bibliographystyle{unsrt}  
\bibliography{main}  %%% Remove comment to use the external .bib file (using bibtex).
\newpage
\clearpage

\section{Supplementary Materials}

\begin{table}[h!]
\centering
\small
\caption{Per-dataset Friedman ranks for the H$_1$ homology on raw Wasserstein$_1$ distance with truncation $2\times r_{enc}$. Lower rank indicates a more faithful preservation of the persistence diagram. Bolded values are per-dataset winner. The bottom row is the mean rank across datasets}
\label{tab:real_topology_per_dataset_rank_2renc}
\begin{tabular}{lccccccccc}
\hline\hline
\textbf{Dataset} & \textbf{BoundaryTPS} & \textbf{CNN} & \textbf{TPS} & \textbf{IB3} & \textbf{CNN+ENN} & \textbf{DROP3} & \textbf{KMeans} & \textbf{SPOTGreedy} & \textbf{ICF} \\
\hline
\textbf{BloodTransfusion} & 5.0 & \textbf{1.0} & 3.0 & 2.0 & 4.0 & 6.0 & 9.0 & 7.0 & 8.0 \\
\textbf{BreastCancer} & 2.0 & \textbf{1.0} & 6.0 & 4.0 & 3.0 & 5.0 & 7.0 & 8.0 & 9.0 \\
\textbf{CardiotocographyA} & \textbf{1.0} & 7.0 & 2.0 & 8.0 & 9.0 & 6.0 & 4.0 & 5.0 & 3.0 \\
\textbf{Diabetes} & 3.0 & \textbf{1.0} & 6.0 & 2.0 & 4.0 & 5.0 & 7.0 & 8.0 & 9.0 \\
\textbf{Digits} & \textbf{1.0} & 3.0 & 2.0 & 4.0 & 8.0 & 7.0 & 6.0 & 5.0 & 9.0 \\
\textbf{Ionosphere} & \textbf{1.0} & 5.0 & 2.0 & 6.0 & 3.0 & 4.0 & 8.0 & 7.0 & 9.0 \\
\textbf{Iris} & 2.0 & 7.0 & 4.0 & 6.0 & 3.0 & 5.0 & 9.0 & 8.0 & \textbf{1.0} \\
\textbf{KCvsKP} & 2.0 & 3.0 & \textbf{1.0} & 5.0 & 6.0 & 4.0 & 9.0 & 8.0 & 7.0 \\
\textbf{MiceProtein} & \textbf{1.0} & 5.0 & 2.0 & 7.0 & 8.0 & 9.0 & 3.0 & 4.0 & 6.0 \\
\textbf{Ozone} & 3.0 & 2.0 & 6.0 & \textbf{1.0} & 4.0 & 5.0 & 8.0 & 9.0 & 7.0 \\
\textbf{Satimage} & 2.0 & \textbf{1.0} & 3.0 & 4.0 & 7.0 & 6.0 & 8.0 & 9.0 & 5.0 \\
\textbf{Sonar} & \textbf{1.0} & 2.0 & 4.0 & 3.0 & 7.0 & 5.0 & 6.0 & 9.0 & 8.0 \\
\textbf{Spambase} & 2.0 & \textbf{1.0} & 5.0 & 3.0 & 4.0 & 6.0 & 8.0 & 7.0 & 9.0 \\
\textbf{Wilt} & 3.0 & 2.0 & 4.0 & \textbf{1.0} & 5.0 & 6.0 & 9.0 & 8.0 & 7.0 \\
\textbf{Wine} & \textbf{1.0} & 7.0 & 4.0 & 8.0 & 6.0 & 5.0 & 2.0 & 3.0 & 9.0 \\
\hline
\textbf{Mean Rank} & 2.00 & 3.20 & 3.60 & 4.27 & 5.40 & 5.60 & 6.87 & 7.00 & 7.07 \\
\hline\hline
\end{tabular}
\end{table}

% ============================================================================

\begin{table}[h!]
\centering
\small
\caption{Per-method topology-preservation performance across all five metrics. The top number in each cell is the Friedman rank. Lower values are more faithful preservation of PD structure; a $\dagger$ marks methods whose rank exceeds the column-best by more than the Nemenyi critical difference at $\alpha = 0.05$. Calculations were done with $2 \times r_{\mathrm{enc}}(X_{\mathrm{train}})$. Rows sorted by H$_1$ W$_1$ rank. Friedman $p$ for each column is in the bottom row.}
\label{tab:real_topology_friedman_summary_2renc}
\begin{tabular}{lccccc}
\hline\hline
\textbf{Method} & \textbf{H$_1$ W$_1$} & \textbf{H$_1$ W$_2$} & \textbf{H$_0$ W$_1$} & \textbf{H$_0$ W$_2$} & \textbf{H$_1$ B} \\
\hline
\textbf{BoundaryTPS} & 2.00 & 2.87 & 5.07 & 6.27$^{\dagger}$ & 5.03 \\
\textbf{CNN} & 3.20 & 2.93 & 2.33 & 2.20 & 3.70 \\
\textbf{TPS} & 3.60 & 4.20 & 4.60 & 5.87$^{\dagger}$ & 5.27 \\
\textbf{IB3} & 4.27 & 4.07 & 3.60 & 3.53 & 3.80 \\
\textbf{DROP3} & 5.60$^{\dagger}$ & 5.33 & 5.67$^{\dagger}$ & 5.73$^{\dagger}$ & 4.93 \\
\textbf{CNN+ENN} & 5.40$^{\dagger}$ & 5.47 & 5.93$^{\dagger}$ & 5.73$^{\dagger}$ & 5.07 \\
\textbf{ICF} & 7.07$^{\dagger}$ & 6.93$^{\dagger}$ & 5.00 & 4.60 & 5.57 \\
\textbf{SPOTGreedy} & 7.00$^{\dagger}$ & 6.73$^{\dagger}$ & 6.00$^{\dagger}$ & 5.40$^{\dagger}$ & 6.03 \\
\textbf{KMeans} & 6.87$^{\dagger}$ & 6.47$^{\dagger}$ & 6.80$^{\dagger}$ & 5.67$^{\dagger}$ & 5.60 \\
\hline
\textbf{Friedman }$p$ & 6.8e-09 & 4.7e-06 & 2.5e-04 & 4.1e-04 & 2.2e-01 \\
\hline\hline
\end{tabular}
\end{table}

\end{document}